\documentclass[conference]{IEEEtran}
\IEEEoverridecommandlockouts
\usepackage{hyperref}
\usepackage{cite}
\usepackage{amsmath,amssymb,amsfonts}
\usepackage{algorithmic}
\usepackage{graphicx}
\usepackage{textcomp}
\usepackage{xcolor}
\def\BibTeX{{\rm B\kern-.05em{\sc i\kern-.025em b}\kern-.08em
    T\kern-.1667em\lower.7ex\hbox{E}\kern-.125emX}}

\newcommand{\paragraphb}[1]{\noindent{\bf #1} }

\DeclareMathOperator*{\argmax}{argmax}

\usepackage{subfigmat}

\usepackage{multirow}
\usepackage{tikz}
\newcommand*\circledA[1]{\tikz[baseline=(char.base)]{
            \node[shape=circle,draw,inner sep=.3pt, fill=red] (char) {#1};}}
\newcommand*\circledB[1]{\tikz[baseline=(char.base)]{
            \node[shape=circle,draw,inner sep=.3pt, fill=gray] (char) {#1};}}
\newcommand*\circledC[1]{\tikz[baseline=(char.base)]{
            \node[shape=circle,draw,inner sep=.3pt, fill=yellow] (char) {#1};}}
\newcommand*\circledD[1]{\tikz[baseline=(char.base)]{
            \node[shape=circle,draw,inner sep=.3pt, fill=cyan] (char) {#1};}}

\newcommand*\circledE[1]{\tikz[baseline=(char.base)]{
            \node[shape=circle,draw,inner sep=.3pt, fill=magenta] (char) {#1};}}

\usepackage{enumitem}
\setitemize{noitemsep,topsep=0pt,parsep=0pt,partopsep=-10pt}

\begin{document}

\title{Fake or Compromised? Making Sense of \\Malicious Clients in Federated Learning}

\author{\IEEEauthorblockN{Hamid Mozaffari}
\IEEEauthorblockA{\textit{University of Massachusetts Amherst}\\
hamid@cs.umass.edu}
\and
\IEEEauthorblockN{Sunav Choudhary}
\IEEEauthorblockA{\textit{Adobe Research}\\
schoudha@adobe.com}
\and
\IEEEauthorblockN{Amir Houmansadr}
\IEEEauthorblockA{\textit{University of Massachusetts Amherst}\\
amir@cs.umass.edu}
}

\maketitle

\begin{abstract}
  Federated learning (FL) is a distributed machine learning paradigm that enables training models on decentralized data. The field of FL security against poisoning attacks is plagued with confusion due to the proliferation of research that makes different assumptions about the capabilities of adversaries and the adversary models they operate under. 
Our work aims to clarify this confusion by presenting a comprehensive analysis of the various poisoning attacks and defensive aggregation rules (AGRs) proposed in the literature, and connecting them under a common framework.
To connect existing adversary models, we present a hybrid adversary model, which lies in the middle of the spectrum of adversaries, where the adversary compromises a few clients, trains a generative (e.g., DDPM) model with their compromised samples, and generates new synthetic data to solve an optimization for a stronger (e.g., cheaper, more practical) attack against different robust aggregation rules.
By presenting the spectrum of FL adversaries, we aim to provide practitioners and researchers with a clear understanding of the different types of threats they need to consider when designing FL systems, and identify areas where further research is needed.
\end{abstract}


\section{Introduction}
Federated learning (FL) is a machine learning paradigm that enables training models on decentralized data, such as mobile devices or edge devices. In FL, each \emph{client} updates the global model using their local data, and communicate the updated model to the  \emph{central server}. Finally, the server  aggregates the updates from all clients using an \emph{aggregation rule} (AGR),  creating the next version of the global model. This approach allows for the training of models on large-scale, non-iid data without collecting clients' original data.

\paragraphb{Fake or compromised? A fork in the literature!} FL is susceptible to \emph{poisoning} by malicious clients who aim to hamper the accuracy of the global model by contributing malicious updates during FL's training process. Based on how the adversary introduces malicious clients in the FL ecosystem, existing works on FL poisoning can be categorized into two major lines of work: 1)  a small percentage  ($<$1\%) of ``actual''  clients are \emph{compromised} by an adversary, e.g., by taking control of some compromised mobile devices; 2) a large percentage ($>$10\%) of \emph{fake}  clients are created and injected into the FL ecosystem, e.g., by creating Sybil accounts or using botnets. The ``compromised'' category~\cite{baruch2019a,shejwalkar2021manipulating, mozaffari2021frl}  targets sophisticated, large-scale applications such as Gboard and Siri that have deployed proper protections against Sybil attacks and botnets. However, these attacks require compromising actual FL devices, which is costly in practice. 

On the other hand, the ``fake'' category~\cite{cao2022mpaf, lin2019free, fraboni2021free} assumes that the adversary can inject large numbers of fake clients, such as spam bots, into the FL ecosystem. Such (large-scale) fake clients cannot be injected into sophisticated applications such as Gboard and Siri as thoroughly discussed by~\cite{shejwalkar2021back}; however, FL applications built on third-party code/software may be vulnerable to such fake clients. 

These two major lines of study are significantly different in terms of their assumptions about the adversary, the adversary model, and the practical settings they represent. This stark disconnect between the adversary model assumptions in the current work leads to confusion about the applicability of a given adversary model to the FL setting of interest.

\begin{figure*} [hbt!]
    \centering
    \includegraphics[width=1.2\columnwidth]{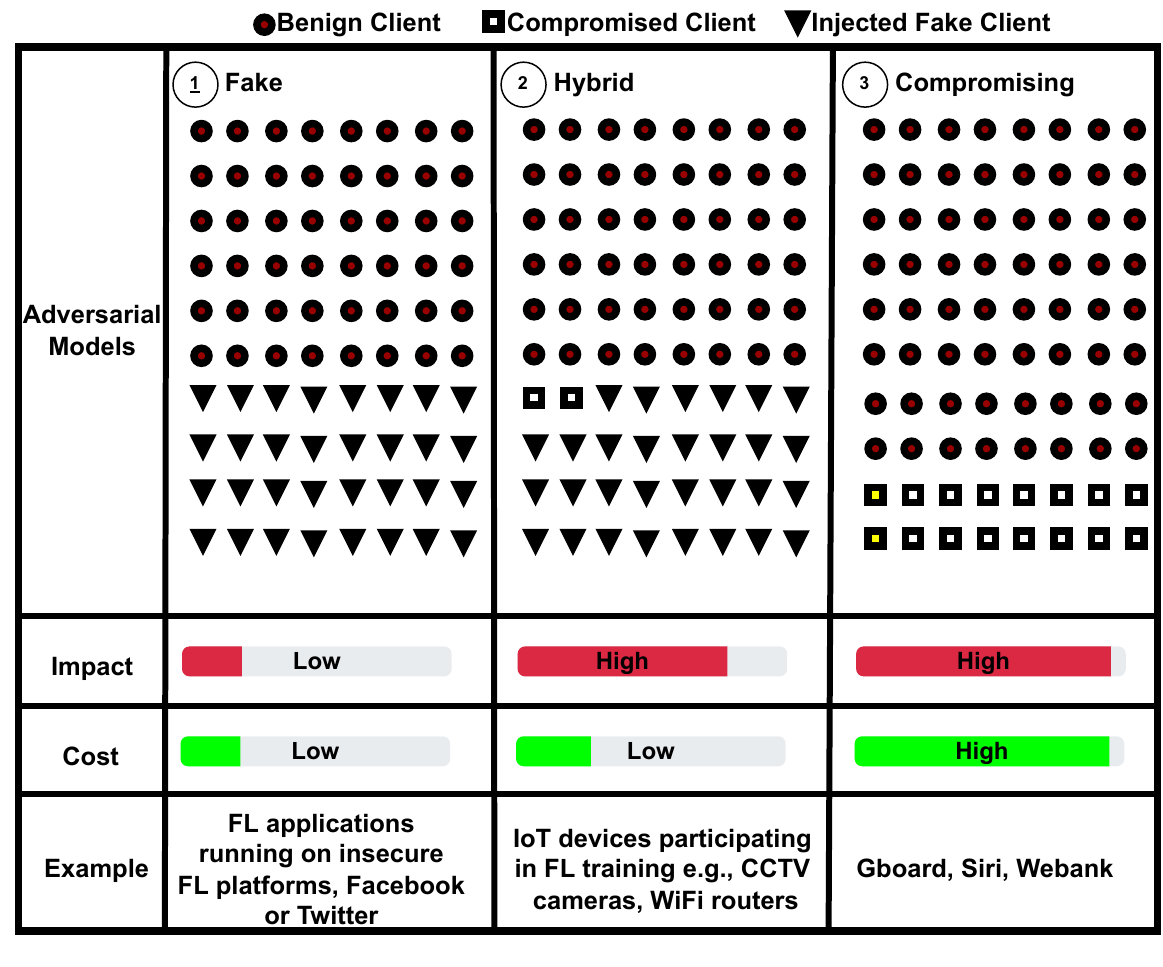}
    \caption{Spectrum of the adversarial models that vary in the number of compromised clients and the number of fake clients injected into the FL system. \textbf{(1)} A scenario of fake clients may occur when FL applications are running on insecure FL platforms, or if we learn an FL model on Facebook or Twitter users, which can have a large number of fake accounts. In this scenario, the adversary can easily introduce fake clients, such as spam bots, into the FL ecosystem; these fake clients do not have any real data and can manipulate the updates they send to the central server. \textbf{(2)} A scenario of hybrid attack may occur when IoT devices participating in FL training, such as CCTV cameras or WiFi routers. An adversary can buy zombies from botnets for compromised and fake clients (more details in Section~\ref{sec:cost}). \textbf{(3)} A scenario of compromised clients may occur in FL applications such as Google's Gboard, Apple's Siri, and Webank. In this scenario, the adversary may use sophisticated techniques such as social engineering, malware injection, or exploiting software vulnerabilities to compromise a small percentage of clients.}
    \label{fig:spectrum}
\end{figure*}

\paragraphb{Presenting a spectrum of adversaries!} 
As opposed to considering two (extreme) adversary models, i.e., compromised and fake, in this paper, we fill the gap between these two adversary models and present a spectrum of adversary models, as sketched in Figure~\ref{fig:spectrum}. We believe that this is essential to
truly understand the poisoning threat to various types of FL deployments in the real world. 
We provide a comprehensive analysis of the different types of poisoning attacks and defensive AGRs that have been proposed in the literature, and we show how these different works can be connected by a common framework.

\paragraphb{Introducing a hybrid adversary model.} 
As discussed above, the literature has only evaluated against two extreme adversary  adversary models, i.e., all compromised and all fake adversarial clients.  
We make the case for a \emph{hybrid adversary model} in which 
the adversary  compromises a very small number of actual users,  and then uses their data to  fabricate a  large number of fake clients (who are supposed to be more impactful than oblivious fake clients considered in the literature). 
Given the quick and broad adoption of FL in various applications, we believe that such hybrid adversary model can be representative of a very large fraction of FL applications in the future. 

Under such a hybrid adversary model, we propose a novel model poisoning attack, called \emph{hybrid attack}, that first leverages the data of compromised clients to \emph{generate} more data using state-of-the-art generative models, e.g., the denoising diffusion probabilistic model (DDPM)~\cite{NEURIPS2020_4c5bcfec, pmlr-v139-nichol21a, kingma2021on, Choi2022PerceptionPT, Karras2022ElucidatingTD, Song2021DenoisingDI, chen2022analog, Qiao2019WeightS, Salimans2022ProgressiveDF, Ho2022ClassifierFreeDG, Sunkara2022NoMS, Jabri2022ScalableAC}. The adversary then uses existing state-of-the-art model poisoning attacks to fabricate poisoned model updates for its compromised and fake clients (which are sent to the FL server).
DDPM is a generative model that has recently gained attention for its ability to learn the underlying structure of complex data distributions from limited and noisy observations. DDPM is based on the idea of diffusion, which is a process of iterative exchange of information between the data points in order to reveal their underlying structure. Specifically, DDPM uses a diffusion process to transform given input data into a latent representation, which captures the underlying structure of the data. This latent representation can then be used to generate new samples that are similar to the original input data.

One key advantage of DDPM is that it is able to learn the structure of the data distribution from a small number of observations, even in the presence of noise. This makes it particularly useful for applications where the data is limited or noisy, such as in the case of compromised clients in federated learning. By using DDPM to generate new samples from a small number of compromised clients, an adversary is able to craft a malicious update for FL poisoning that is representative of the data distribution of the benign clients.

In addition to our focus on DDPM for generating synthetic data in poisoning attacks, we also explore another potent alternative which is utilizing Deep Convolutional Generative Adversarial Networks (DCGAN)~\cite{radford2015unsupervised}. Our exploration into DCGAN, known for its proficiency in generating high-fidelity images, reveals its significant effectiveness in creating quality data for use in poisoning attacks on global models. Comparative analysis shows that while DCGAN and DDPM are both highly effective, DDPM demonstrates a superiority, particularly in generating more realistic synthetic data from limited and noisy datasets. This distinction is crucial in our hybrid attack model, where the quality of the generated data plays a pivotal role in determining the overall impact and effectiveness of the attack.

\noindent\textbf{Empirical evaluations:}
We provide extensive evaluations of existing attacks as well as our hybrid attacks under various adversary models obtained by combining the spectrum of adversaries and defenders discussed above. 
We experiment with two datasets, FEMNIST and CIFAR10, in real-world heterogeneous FL settings.

\noindent In summary, our key contributions are as follows:
\begin{itemize}
    \item  The literature of FL poisoning has forked into two separate lines of work that assume two differing adversary models, i.e.,  fake and compromised, as introduced earlier. 
    Our work aims to highlight the differences between these two lines of work by contrasting their application scenarios, assumptions, and costs. 
    
    \item We fill the gap between fake and compromised adversary models by introducing a spectrum of adversary models, which we call hybrid. Through extensive experiments we  demonstrate how the hybrid adversary models establish trade-offs between attack accuracy and attack cost in comparison to the fake and compromised models. 
    
    \item We design and evaluate   novel FL poisoning attacks that work under  the newly introduced hybrid adversary model. 
    Our attack leverages denoising diffusion probabilistic model (DDPM) to generate poisoning data for fake clients based on the data collected from a small number of compromised clients. 

    \item We explore the impact of using DCGAN~\cite{radford2015unsupervised} alongside DDPM, demonstrating DDPM's superior effectiveness in creating more damaging adversary updates.


\end{itemize}

 We hope that our work can bridge the gap between existing works on FL poisoning,  providing practitioners and researchers with a clear understanding of the various adversary models that need to be considered when designing FL systems.

\section{Background}

\subsection{Federated learning (FL)}\label{background:fl}
In FL~\cite{mcmahan2017communication, kairouz2019advances, konevcny2016federated}, 
$N$ clients collaborate to train a global model  without directly sharing their data. 
In  round $t$, the service provider (server) selects $n$ out of $N$ total clients and  sends them the most recent global model $\theta^t$. Each client trains a local model for $E$ local epochs on their data starting from the $\theta^t$ using stochastic gradient descent (SGD). Then the client  sends back the calculated gradients ($\theta_k^t$ for $k$th client) to the server. The server then aggregates the collected gradients and  updates the global model for the next round.
In the untargeted attack~\cite{baruch2019a, fang2020local,shejwalkar2021manipulating}, the goal is to reduce the utility for all (or most)  test inputs. In this work, we focus on untargeted poisoning due to its severity in practice~\cite{shejwalkar2021back}.

In order to make FL robust against such malicious clients, the literature has designed various \emph{robust aggregation rules (AGR)}~\cite{blanchard2017machine, DBLP:conf/icml/MhamdiGR18, YinCRB18, mozaffari2021frl, chang2019cronus, mozaffari2022fedperm}, which aim to remove or attenuate the updates that are more likely to be malicious according to some criterion. 
Unfortunately, these robust AGRs are not very effective in  non-convex FL settings, and multiple works have demonstrated strong targeted~\cite{wang2020attack,bhagoji2019analyzing} and untargeted attacks~\cite{shejwalkar2021manipulating,fang2020local} on them.

\subsection{Diffusion models}\label{background:diffusion}
The \emph{denoising diffusion probabilistic model} (DDPM)~\cite{NEURIPS2020_4c5bcfec, pmlr-v139-nichol21a, kingma2021on, Choi2022PerceptionPT, Karras2022ElucidatingTD, Song2021DenoisingDI, chen2022analog, Qiao2019WeightS, Salimans2022ProgressiveDF, Ho2022ClassifierFreeDG, Sunkara2022NoMS, Jabri2022ScalableAC} is a generative model that aims to learn the underlying structure of a complex data distribution from a small number of noisy observations. DDPM is based on the idea of diffusion, which is a process of iteratively exchanging information between the data points in order to reveal their underlying structure.

To learn the structure of the data distribution, DDPM uses a diffusion process to transform the given input data into a latent representation. This latent representation is obtained through a series of diffusive steps, which are defined by a diffusion operator. At each diffusive step, the data points are transformed by exchanging information with their neighbors in the data space. This process is repeated until the latent representation converges to a stable state, which captures the underlying structure of the data. Once the latent representation is obtained, it can be used to generate new samples that are similar to the original input data.

In summary, the ability of DDPM to learn the structure of complex data distributions from limited and noisy observations makes it a valuable tool for researchers seeking to generate new samples for various purposes. In the context of FL poisoning, DDPM can be used to generate new samples from a small number of compromised clients, enabling an adversary to craft a more effective malicious update.
\section{Types of Byzantine-robust aggregation rules} \label{sec:AGRs}
The existing Byzantine-robust aggregation rules (AGRs) for federated learning can be categorized into three categories: non-robust AGRs, AGRs agnostic to poisoning attacks, and AGRs that adapt to or are aware of the poisoning attacks in FL ecosystem.



\subsection{Non-robust AGR} 
Non-robust aggregation rules, such as federated averaging (FedAvg)~\cite{konevcny2016federated, mcmahan2017communication}, do not consider the presence of malicious clients in the federated learning ecosystem. Therefore, such AGRs simply aggregate the model updates received from all clients by computing a non-robust function of the updates. While these approaches are generally simpler and easy to implement, they are vulnerable to model and data poisoning attacks~\cite{shejwalkar2021back,shejwalkar2021manipulating,fang2020local,mozaffari2021frl}. Examples of such non-robust AGRs include FedAvg~\cite{konevcny2016federated, mcmahan2017communication}, SCAFFOLD~\cite{karimireddy2020scaffold}, and MIME~\cite{karimireddy2020mime}. Below we introduce FedAVG.

\textit{FedAVG:}\label{indiff_fedavg} In non-adversarial FL settings, i.e., without any malicious clients, the dimension-wise Average (FedAvg)~\cite{konevcny2016federated, mcmahan2017communication}  is an effective AGR. In fact, due to its efficiency, Average is the only AGR implemented by FL applications in practice~\cite{DBLP:journals/corr/abs-2007-10987, DBLP:journals/corr/abs-2102-08503}. 
However, even if there is a single malicious client, it can destroy the global model by sharing very large local model updates~\cite{blanchard2017machine}.

\subsection{Robust AGRs agnostic to FL poisoning}
Robust AGRs, such as Median and Norm-Bounding, are robust in that they aim to reduce the impact of malicious clients' updates. But, they are \emph{agnostic} in that they do not have any knowledge of the specifics of the attacks, e.g., they do not know the number of malicious updates in each round. These rules use techniques from robust statistics, such as outlier removal or clipping the norms of updates, to exclude or mitigate the impact of malicious updates during the aggregation process. 
While this can provide some protection against model poisoning attacks, it may not be sufficient if the number of malicious updates is large or if the malicious updates are able to evade detection.

\subsubsection{Median}\label{agn_aware_median} 
Median is a well-known robust statistic that is less sensitive to outlier values.
To compute the coordinate-wise Median~\cite{YinCRB18} of updates, for each of the coordinates of the update, we compute the median of all values from all client updates. Median AGR is particularly useful in situations where there may be a small number of malicious or outlier clients that provide model updates that are significantly different from the others. By using Median, these outlier model updates will not have as much influence on the global model as they would with non-robust AGRs, such as FedAvg. 

\subsubsection{Norm-Bounding}\label{agn_aware_normb} 
This AGR~\cite{sun2019can} bounds the L2 norm
of all submitted client updates to a fixed threshold $\tau$, with the intuition that the effective poisoned updates should have high norms. For a threshold  $\tau$ and an update $\triangledown$, if the norm, $||\triangledown||_2 > \tau$, $\triangledown$ is scaled by $\frac{\tau}{||\triangledown||_2}$, otherwise, the update is not
changed. The final aggregate is an average of all the updates, scaled or otherwise.

\subsection{Robust AGRs that adapt to FL poisoning}
Adaptive aggregation rules have the advantage of knowing the number of malicious updates in each round for aggregation. These rules use this information to adapt their aggregation process in order to mitigate the impact of malicious updates on the final model. While this can be an effective way to protect against model poisoning attacks, it requires a way to accurately estimate the number of malicious updates, which can be a challenging task.

\subsubsection{Multi-Krum}\label{ada_aware_mkrum} Blanchard et al.~\cite{blanchard2017machine} proposed Multi-Krum
AGR as a modification to their own Krum AGR. Multi-Krum selects an update using Krum and adds it to a selection set, $S$. Multi-Krum repeats this for the remaining updates (which remain after removing the update that Krum selects) until $S$ has $c$ updates such that $n-c>2m+2$, where $n$ is the number of selected clients and $m$ is the number of compromised clients in a given round. Finally, Multi-Krum averages the updates in $S$.

\subsubsection{Trimmed-Mean}\label{ada_aware_trmean} Yin et al.~\cite{YinCRB18} proposed Trimmed-Mean that aggregates each dimension of input updates separately. It sorts the values of the $j^{th}$-dimension of all updates. Then it removes $m$ (i.e., the number of compromised clients) of the largest and smallest values of that dimension, and computes the average of the rest of the values as its aggregate for the dimension $j$.

\section{Distinguishing Fake And Compromised Adversary Models}
A poisoning attack is either \emph{data} or \emph{model} poisoning attack: in data poisoning, the adversary can poison only the data on malicious client device, while in model poisoning, the adversary can directly manipulate/poison the model updates of the malicious clients. In this work, we focus on model poisoning, as it is strictly stronger than data poisoning~\cite{shejwalkar2021back,shejwalkar2021manipulating}; hence, poisoning in any context refers to model poisoning, unless stated otherwise.

In this section, we discuss the types of adversaries in FL
that mount attacks with varying costs and impacts.
Fake clients are utilized in less secure FL systems where client injection is relatively easy. They are cost-effective but limited in their impact. Compromised clients appear on more secure FL systems, requiring sophisticated methods such as malware to take control of actual clients. Our novel contribution in the next section (Section~\ref{sec:hybridattack}) is the introduction of a hybrid adversary model that combines these two approaches. This model reflects a more realistic scenario where an attacker might use a mix of fake and compromised clients, leveraging the ease of injecting fake clients and the authenticity of compromised ones to mount a more effective attack. This hybrid model embodies the complexity of real-world threats and underlines the need for versatile defensive strategies in FL.



\subsection{Adversary with fake clients}
In federated learning (FL) systems, an attacker can inject fake clients in order to send arbitrary fake local model updates to the cloud server. This type of attack is more affordable and easier to perform than compromising genuine clients, as the attacker does not need to bypass anti-malware software or evade anomaly detection on the clients' devices. Instead, the attacker can emulate fake clients using open source projects or free software such as android emulators, which can be run on a single machine to emulate multiple instances, i.e., multiple FL clients, significantly reducing the attack cost. Fake clients also offer the advantage of being fully controlled by the attacker, as Android emulators can grant root access to the devices. These factors make model poisoning attacks using fake clients a realistic threat in FL systems.

Cao et al.\ proposed MPAF~\cite{cao2022mpaf}, a method of attacking FL systems through the injection of fake clients. In MPAF, the attacker selects a randomly initialized model as the base model ($\theta'$), whose test accuracy is close to random guessing, and crafts fake local model updates to force the global model to mimic the base model. This is done by subtracting the current global model parameters ($\theta^{t}$ for the FL round $t$) from the base model parameters and scaling the fake local model updates by a factor $\lambda$ to amplify their impact. Equation~\ref{eq:Mpaf} shows the malicious updates of the fake clients.

\begin{equation}\label{eq:Mpaf}
\begin{aligned}
    \theta_{m \in [M]}^t = \lambda (\theta' - \theta^{t})
\end{aligned}
\end{equation}
\noindent where $\theta_{m \in [M]}$ are the malicious model updates for $M$ injected fake clients, and $\theta'$ is the randomly initialized  base model.

To perform MPAF, the attacker must have minimum knowledge of the FL system, which means that they only have access to the global models received during training. Despite this limited information, MPAF is able to effectively manipulate the global model by driving it towards the base model in each FL round. This is done by calculating fake local model updates ($\theta_{m \in [M]}$), which are then aggregated by the cloud server along with genuine local model updates from genuine clients. The attacker can choose a large $\lambda$ to ensure that the attack is effective even after aggregation.

In our paper, we refer to this attack as the Fake attack. This attack is characterized by the minimal knowledge and ability required from the adversary who controls the fake clients. Specifically, the fake attack is the simplest attack of this kind in FL and represents one end of the spectrum of attacks based on the impact and cost of the attack.

\subsection{Adversary with compromised clients} \label{sec:comp_attacks}
To evaluate the robustness of various FL algorithms, we use state-of-the-art model poisoning attacks from~\cite{shejwalkar2021manipulating}.
The attack proposes a general FL poisoning framework and then tailors it to specific FL settings. First, it computes an average $\theta^{b,t} = f_{\text{avg}} (\theta_{c\in[C]}^{t})$ of benign updates, $\theta_{c\in[C]}^{t}$, available to the adversary in the FL round $t$. Then it perturbs $\theta^{b,t}$ in a \emph{dynamic, data-dependent malicious direction} $\omega$ to calculate the final poisoned update $\theta_{c \in [C]}^{t,m}= \theta^{b,t} + \gamma \omega$. The attack, called \emph{DYN-OPT}, finds the largest $\gamma$ that successfully circumvents the target AGR. DYN-OPT is much stronger than its predecessors, because it finds the largest $\gamma$ and uses a tailored dataset $\omega$. In the following, we detail the DYN-OPT attacks against the AGRs from Section~\ref{sec:AGRs} that we consider in this work.

\subsubsection{FedAVG} DYN-OPT attack against FedAVG is quite straightforward and uses a random direction $\omega$ and a very large value $\gamma$ to compute the poisoned update $\theta_{c \in [C]}^{t,m}$.

\subsubsection{Mutli-Krum}
As described in Section~\ref{ada_aware_mkrum}, Multi-Krum uses Krum iteratively to construct a selection set $S$ and computes the average of the updates in the selection set as its aggregate. Therefore, DYN-OPT aims to maximize the perturbation $\gamma\omega$ used to compute the poison update $\theta_{c \in [C]}^{t,m}$, while ensuring that Multi-Krum selects all its poison updates in $S$. Note that this strategy minimizes the number of benign updates in $S$ and maximizes $\gamma\omega$ by increasing the poisoning impact of malicious updates on the final aggregate. The optimization problem we solve to mount DYN-OPT on Multi-Krum is given in~\eqref{dyn_opt_mkrum}.

\begin{align}\label{dyn_opt_mkrum}
    \argmax_{\gamma} \;\; & |\{\theta_{c \in [C]}^{t,m} \in f_{\text{mkrum}} \left( \theta_{c \in [C]}^{t,m} \cup \theta_{i\in[C+1,n]}^t \right) \}|  \;\;\;\; \\ 
    &\text{s.t. }  \;\;\;\; \theta_{c \in [C]}^{t,m} = \theta^{b,t} +\gamma \omega \nonumber
\end{align}

\subsubsection{Trimmed-Mean and Median}
For Trimmed-Mean and Median AGRs, DYN-OPT solves the optimization given in~\eqref{dyn_opt_trmean}. Following~\cite{shejwalkar2021manipulating}, we fix the perturbation $\omega$ and keep all poisoned updates the same. The objective here is to maximize the $L_2$ norm of the distance between the benign update reference $\theta^{b,t}$
and the aggregate, $f_{\text{agr}}(.)$, calculated using $f_{\text{agr}}\in\{f_{\text{trmean}}, f_{\text{median}}\}$ on the set of benign and malicious updates.

\begin{align}\label{dyn_opt_trmean}
    \argmax_{\gamma} \;\; & \Vert \theta^{b,t} -f_{\text{agr}} \left(  \theta_{c \in [C]}^{t,m} \cup \theta_{i\in[C+1,n]}^t \right) \Vert_2    \;\;\;\; \\
    & \text{s.t. }  \;\;\;\; \theta_{c \in [C]}^{t,m} = \theta^{b,t} +\gamma \omega \nonumber
\end{align}

\subsubsection{Norm-Bounding} We formulate the DYN-OPT attack against AGR bound to Norm using the original framework proposed in~\cite{shejwalkar2021manipulating}. More specifically, to circumvent Norm-Bounding, the norm of the poisoned update should be less than the threshold norm, $\tau$, used by Norm-Bounding AGR. Therefore, to compute the poison update $\theta_{c \in [C]}^{t,m}$ using DYN-OPT, we can scale the norm of the original poison update, $\theta^{b,t} + \gamma\omega$, to $\tau$. The final poisoned update would be $\theta_{c \in [C]}^{t,m} = \text{Scale}(\theta^{b,t} + \gamma\omega, \tau)$, where $\text{Scale}(u,\tau) = u\cdot\text{min}(1, \frac{\tau}{\Vert u\Vert_2})$.

\section{Our proposed hybrid adversary model} \label{sec:hybridattack}
Compromising real clients in FL to launch a model poisoning attack can be a challenging task for an attacker. This is because genuine clients participating in FL are typically owned and controlled by different entities (e.g., individual users in cross-device FL and hospitals in cross-silo FL), and the attacker should get access to and take control of these clients in order to manipulate the updates they send to the server.

One way an attacker might try to do this is by using malware or phishing attacks to compromise clients. However, successfully executing these types of attacks requires a certain level of skill and resources, and the attacker would need to be able to bypass any security measures that the clients have in place. Additionally, the cost of compromising a large number of genuine clients can be high, as the attacker would need to pay for access to undetected zombie devices or other resources. This may make it infeasible for the attacker to compromise a large fraction of genuine clients, which is typically necessary for a model poisoning attack to be successful.

\begin{figure} 
    \centering
    \includegraphics[width=1.05\columnwidth]{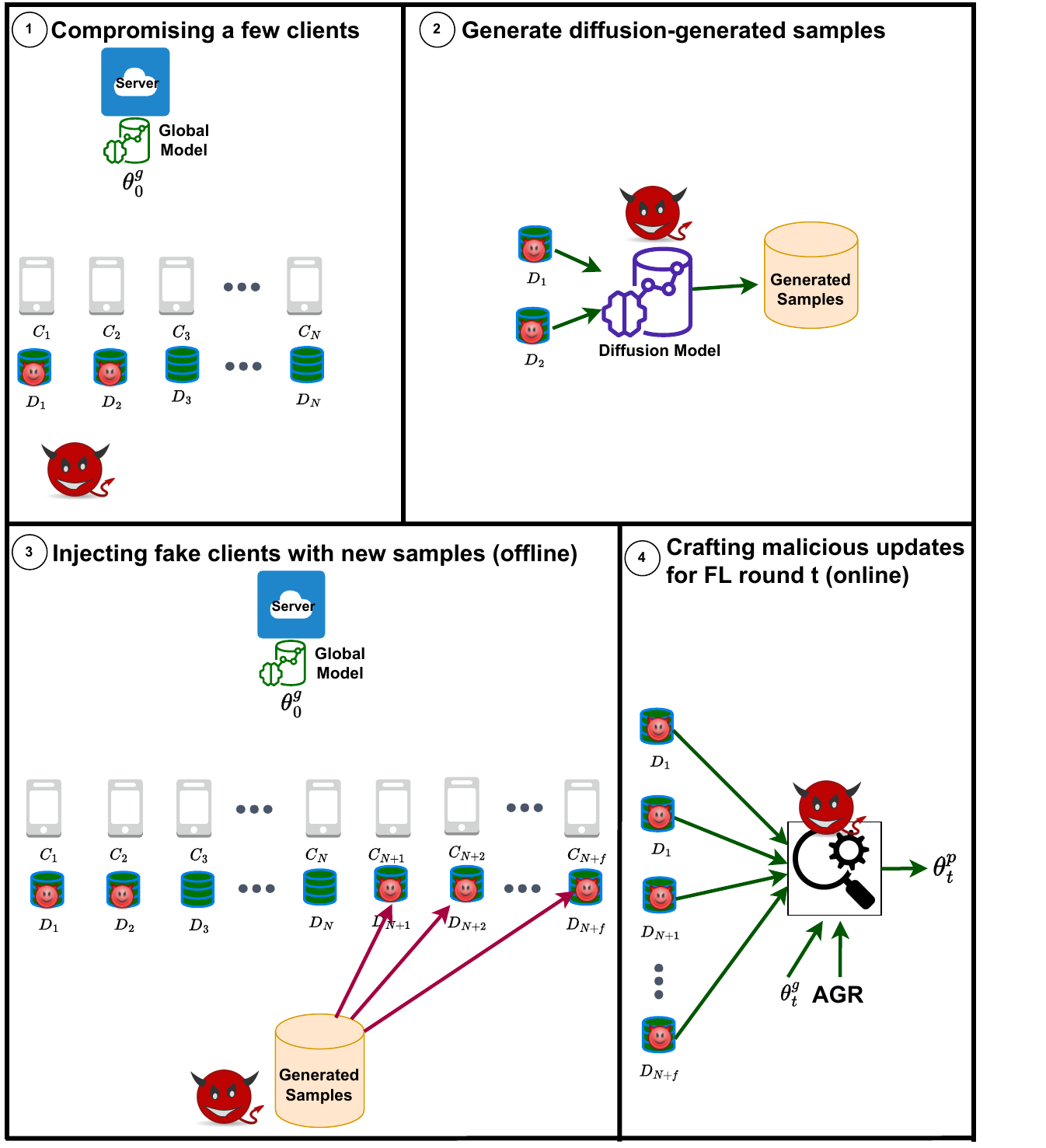}
    \caption{Our novel hybrid attack pipeline: The adversary of hybrid attack lies in the middle of the spectrum of FL poisoning adversaries. The hybrid attack adversary compromises a few real FL clients, trains a denoising diffusion probability model (DDPM) on their real data, and generates new synthetic data to solve an optimization to generate malicious updates to mount strong model poisoning attacks against the target robust aggregation rules. Notably, if high-quality public data that mirrors real client data distribution is available, it can replace the initial data-gathering step in this process, although such data may not be readily available in proprietary contexts. Finally, the adversary shares the malicious update with the FL server via the compromised clients as well as (cheap to inject) fake clients.}
    \label{fig:hybrid_attack}
\end{figure}

Another factor that makes it difficult to compromise real clients in FL is the decentralized nature of the system. In FL, clients are typically distributed across a wide geographical area and may have different levels of security and defenses in place. This can make it difficult for the attacker to gain access to and take control of a large number of clients simultaneously. 

In general, the combination of technical challenges and the high cost of compromising genuine clients in FL makes it a difficult task for an attacker to launch a successful model poisoning attack using only compromised clients. 

Instead, we propose to use both fake and compromised clients to mount a hybrid attack. Figure~\ref{fig:hybrid_attack} shows the pipeline of our hybrid attack: The hybrid adversary first compromises a few real clients and then uses their data to generate synthetic data using a DDPM (Section~\ref{background:diffusion}). Next, the adversary uses these synthetic data to emulate FL clients and uses the model poisoning attacks (Section~\ref{sec:comp_attacks}) to craft strong malicious updates. The injected fake clients and compromised clients submit the generated malicious update if the server selects them in that FL round for their local updates.

Note that in Figure~\ref{fig:hybrid_attack}, Step 1 can be removed if the adversary is able to obtain (high-quality) data samples that represent the data distribution of typical clients. For example, if abundant public data is available related to the target FL task, the adversary can simply use such public data to synthesize the poisoning data for its fake clients. However, high-quality (i.e., representative) public data is not always available, especially in proprietary applications. 

\subsection{Comparing the costs of different adversaries} \label{sec:cost}
In this section, we discuss the cost of the three types of attacks discussed above: fake, hybrid, and compromised. We assume that the cost of compromising a client is $c$ and the cost of creating a fake client is $f$; depending on the scenario, $c$ and $f$ can vary widely, but generally the cost of a fake client is much lower than that of a compromised client, that is, $f \ll c$. 
Furthermore, we assume $\alpha_f$ fake clients in the fake attack, $\beta_c$ compromised clients in the compromised attack, and $\alpha_h$ fake and $\beta_h$ compromised clients in the hybrid attack. 


If the number of malicious clients in the three attacks is the same, that is, $\alpha_f = \alpha_h + \beta_c = \beta_c$, the cost of each of the attacks is as follows: $f\cdot\alpha_f$ for the fake attack, $f\cdot\alpha_h + c\cdot\beta_h$ for the hybrid attack, and $c\cdot\beta_c$ for the compromised attack. Next, note that in our hybrid attack, we use very few compromised clients to launch a very large number of fake clients, i.e., $\alpha_h \gg \beta_h$, which also implies that the number of fake clients in our hybrid attack is very close to that in the fake attack, i.e., $\alpha_h \approx \alpha_f$. Hence, the order of the cost of the three attacks is: $\text{cost}_f < \text{cost}_h \ll \text{cost}_c$, with $\text{cost}_f$ and $\text{cost}_h$ being very close.

Let us consider a concrete scenario involving IoT devices, e.g., CCTV traffic cameras or WiFi routers. The goal of the adversary is to mount a model poisoning attack against an IoT application, e.g., predicting traffic at a certain location. The application stores and uses images from traffic cameras, and trains a global image classification model using FL. With high probability, these IoT devices are also part of some botnet, and the cost of owning such zombie devices in a botnet can be as low as $1$. However, all IoT devices need not have the target application, e.g., many CCTV cameras may not have required software/hardware updates. For concreteness, consider that 1\% of the devices have the target application. Furthermore, note that, generally, the botnet owners do not know what all applications are running on the zombie devices. 

Therefore, in case the compromised attack requires $m$ malicious clients, where the zombie IoT devices must have the application, the adversary will have to buy $100m$ devices to ensure that $m$ of them have the target application and discard $99m$ devices. However, in the case of our hybrid attack, the adversary just needs to ensure that $m'\ll m$ devices have the application (and, therefore, the required data) and should buy max($100m', m$) devices. Then they can install the target application on the $m-m'$ devices and populate them with synthetic data. In the case of a fake attack, the adversary simply has to buy $m$ devices. If the cost of buying a zombie device is $c$, the costs of compromised, hybrid and fake attacks are $100mc \gg 100m'c > mc$; the first inequality holds because $m\gg m'$.
\section{Experimental Setup}\label{sec:setupp}
In this section, we first introduce the evaluation datasets, the model architectures, and the hyperparameter settings. Next, we define our evaluation metrics, and finally, we explain how we generate new data samples using DDPM.

\subsection{Datasets and hyperparameters}
In this work, we conduct experiments on two datasets, CIFAR10~\cite{krizhevsky2009learning} and FEMNIST~\cite{DBLP:journals/corr/abs-1812-01097, DBLP:conf/ijcnn/CohenATS17}, in order to evaluate the performance of different Byzantine robust aggregation under different adversary models.

\paragraphb{CIFAR10} dataset is a widely used image classification dataset consisting of 60,000 32x32 color images in 10 classes, with 6,000 images per class. There are 50,000 training images and 10,000 test images.
For this dataset, we use VGG9 architecture. 
For local training in each FL round, each client uses 5 epochs.
Each client uses SGD with learning rate of 0.01, momentum  of 0.9, weight decay of 1e-4, and batch size of 8. 

\paragraphb{FEMNIST} is a character recognition classification task with 3,400 clients, 62 classes (52 for upper and lower case letters and 10 for digits), and 671,585 gray-scale images. Each client has data of their own handwritten digits or letters. For this dataset, we use LeNet architecture. 
For local training in each FL round, each client uses 2 epochs. 
Each client uses SGD with learning rate of 0.01, momentum of 0.9, weight decay of 1e-4,  and batch size of 10.

\paragraphb{Data distribution:}
Most real-world FL settings have heterogeneous client data, hence following previous works~\cite{reddi2020adaptive, hsu2019measuring}, we distribute CIFAR10 datasets among 1,000 clients in non-iid fashion using \emph{Dirichlet} distribution with parameter $\beta=0.5$. Note that, increasing  $\beta$ results in more iid datasets.
FEMNIST is naturally distributed non-iid among 3,400 clients.

\begin{figure}[hbt!]
    \centering
    \includegraphics[width=0.99\columnwidth]{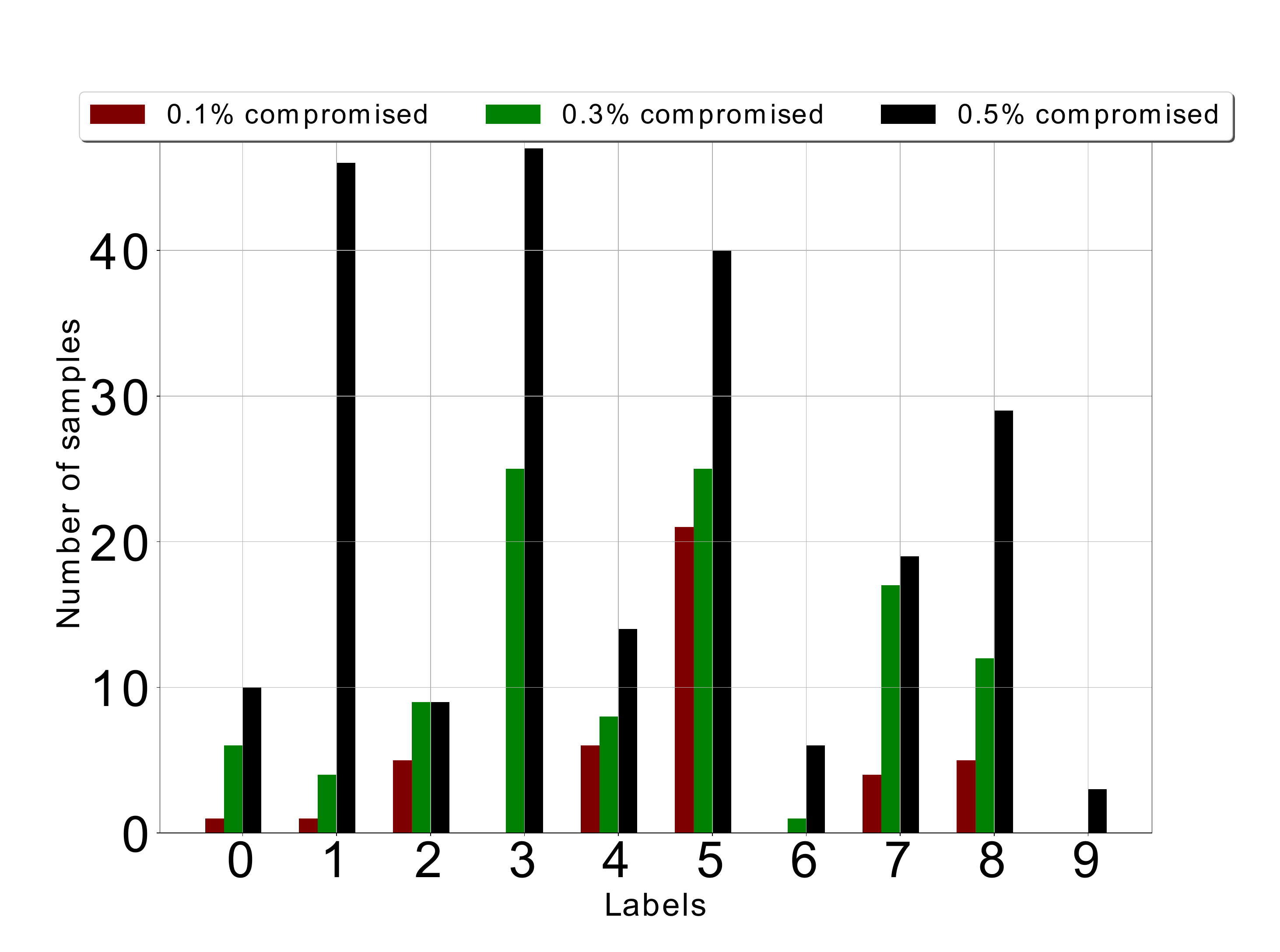}
    \caption{Number of samples for each label when the attacker compromised 0.1\% (1 client), 0.3\% (3 clients), and 0.5\% (5 clients) in our data distribution (fixed through all the experiments) for learning CIFAR10 distributed over 1000 clients.}
    \label{fig:Number_data_samples_CIFAR10_collection}
\end{figure}

\subsection{Evaluation metric}
We run all the experiments for 2000 global rounds of FL for CIFAR10, and 1000 global rounds for FEMNIST, while selecting 25 clients in each round randomly. At the end of each FL round, we calculate the test accuracy of the global model on the test data, and update the maximum test accuracy. We run each experiment with 5 different random seeds, and we report the median and standard deviation of the maximum test accuracies in our experiments.

\paragraphb{Attack impact metric ($I_{\theta}$):} We define attack impact, $I_{\theta}=A_\theta-A_{\theta}^{M}$, as the reduction of the accuracy of the global model when the attack is launched.  ($A_\theta$) denotes the maximum accuracy that the global model achieves overall FL training rounds without the presence of any malicious clients. $A_{\theta}^{M}$ for an attack shows the maximum accuracy of the model under a given attack. 
In our Tables, we report both the maximum test accuracies and Attack Impacts.

\paragraphb{Attack Cost:} Analyzing the cost efficiency tradeoffs of different poisoning attacks in federated learning is crucial for understanding the severity and impact of such attacks. In Section~\ref{sec:cost}, we present a cost analysis of various poisoning attacks across the spectrum of adversary models, ranging from fake to compromising attacks. We assume that an adversary can acquire control of zombie devices in a botnet for \$1 per device. Furthermore, we consider that only 1\% of these devices possess the target application with real data. This implies that out of 100 purchased zombie devices, 99 do not have any real data (suitable for fake attacks), while one device has access to real data, which can be used for a compromising attack.

In a compromising attack scenario that necessitates $m$ malicious clients, with the requirement that the zombie IoT devices have the target application, the attacker must acquire $100m$ devices to confirm that $m$ devices possess the target application while discarding the other $99m$ devices. On the other hand, in our proposed hybrid attack, the attacker only needs to make certain that $m' \ll m$ devices contain the application (along with the required data) and purchase max($100m', m$) devices. They can then install the target application on $m-m'$ devices and populate them with artificial data. In the case of a fake attack, the attacker simply needs to obtain $m$ devices.

For instance, if the attacker aims to launch a compromising attack with 100 malicious clients, they would need to purchase 10,000 zombie devices. Assuming a cost of \$1 per control of each device, the total cost amounts to \$10,000. In contrast, if the attacker desires 100 fake clients, the cost would be \$100. If the attacker wants a hybrid attack with 3 compromised clients possessing real data and 97 fake clients, the cost would be \$300. However, if the attack requires 1 real client and 99 fake clients for a compromising attack, the cost would be \$100. We provide the cost of each attack scenario in each table based on the required number of malicious clients and the type of attack. 

\begin{figure*}[h]
    \begin{center}
    \begin{subfigmatrix}{3} 
    \centering
      \subfigure[Generated by DDPM using 0.1\% (1 client) compromised]{
      \includegraphics[width=0.31\textwidth]{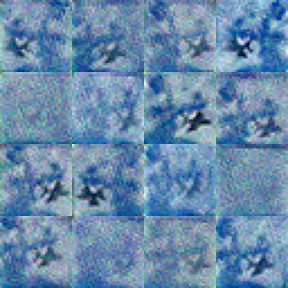}}
    \centering
      \subfigure[Generated by DDPM using 0.3\% (3 clients) compromised]{
      \includegraphics[width=0.31\textwidth]{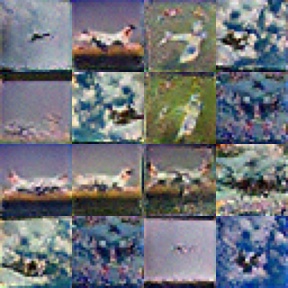}}
    \centering
      \subfigure[Generated by DDPM using 0.5\% (5 clients) compromised]{
      \includegraphics[width=0.31\textwidth]{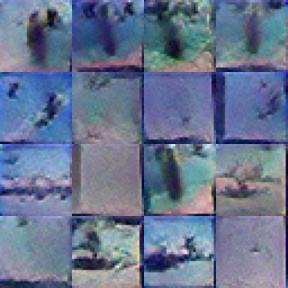}}
    \centering
      \subfigure[Generated by DCGAN using 0.1\% (1 client) compromised]{
      \includegraphics[width=0.31\textwidth]{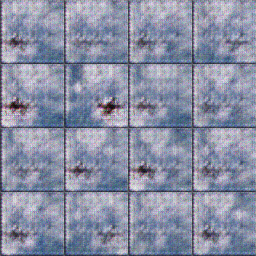}}
    \centering
      \subfigure[Generated by DCGAN using 0.3\% (3 clients) compromised]{
      \includegraphics[width=0.31\textwidth]{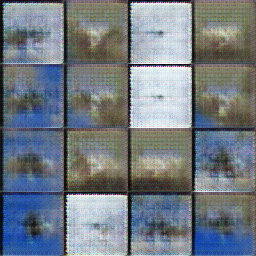}}
    \centering
      \subfigure[Generated by DCGAN using 0.5\% (5 clients) compromised]{
      \includegraphics[width=0.31\textwidth]{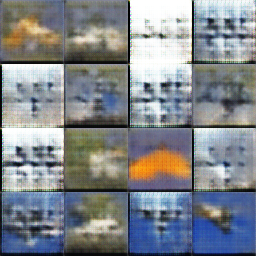}}
    \end{subfigmatrix}
    \end{center}
    \caption{Airplanes generated by DDPM and DCGAN using different percentages of compromised client's data in our hybrid attack.}
    \label{fig:DDPM_airplane}
\end{figure*}

\begin{figure*}[hbt!]
    \begin{center}
    
    \begin{subfigmatrix}{3}
    \centering
      \subfigure[CIFAR10 + Median (No attack acc=76.05\%)]{
      \includegraphics[width=0.32\textwidth]{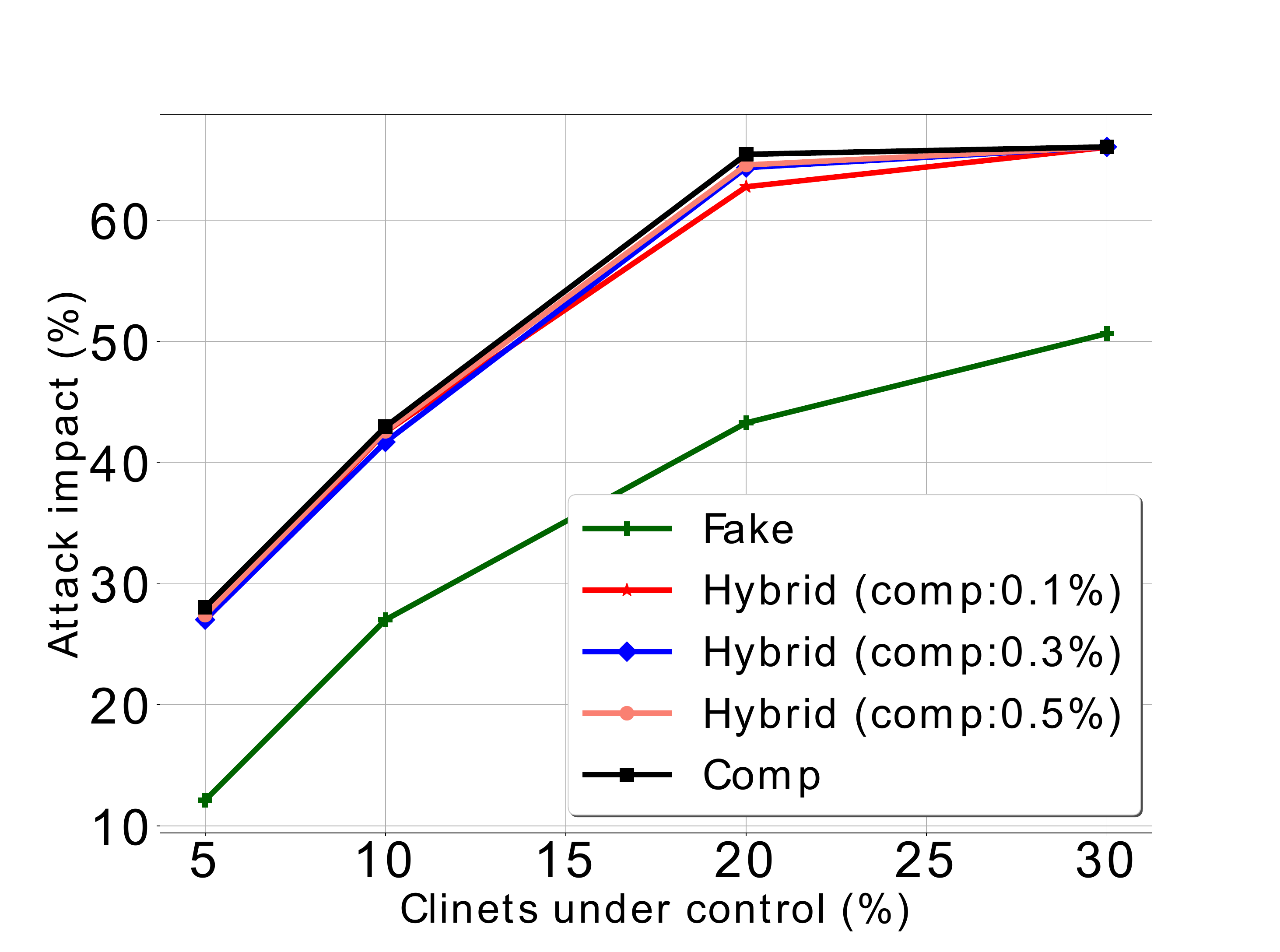}}
     \centering
      \subfigure[CIFAR10 + NB ($\tau=2.0$) (No attack acc=83.68\%)]{
      \includegraphics[width=0.32\textwidth]{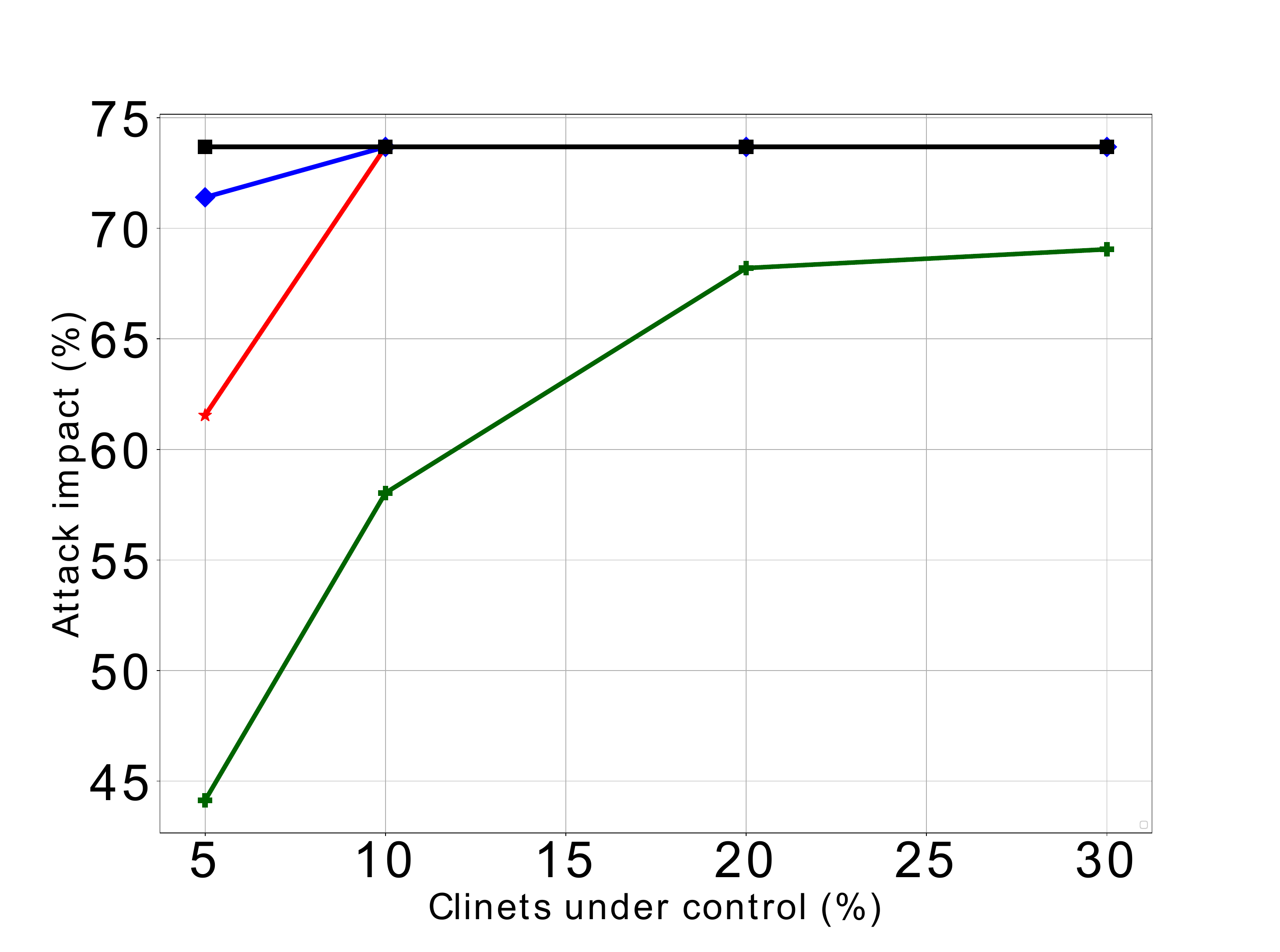}}
      \centering
      \subfigure[CIFAR10 + NB $\tau=0.5$ (No attack acc=78.86\%) ]{
      \includegraphics[width=0.32\textwidth]{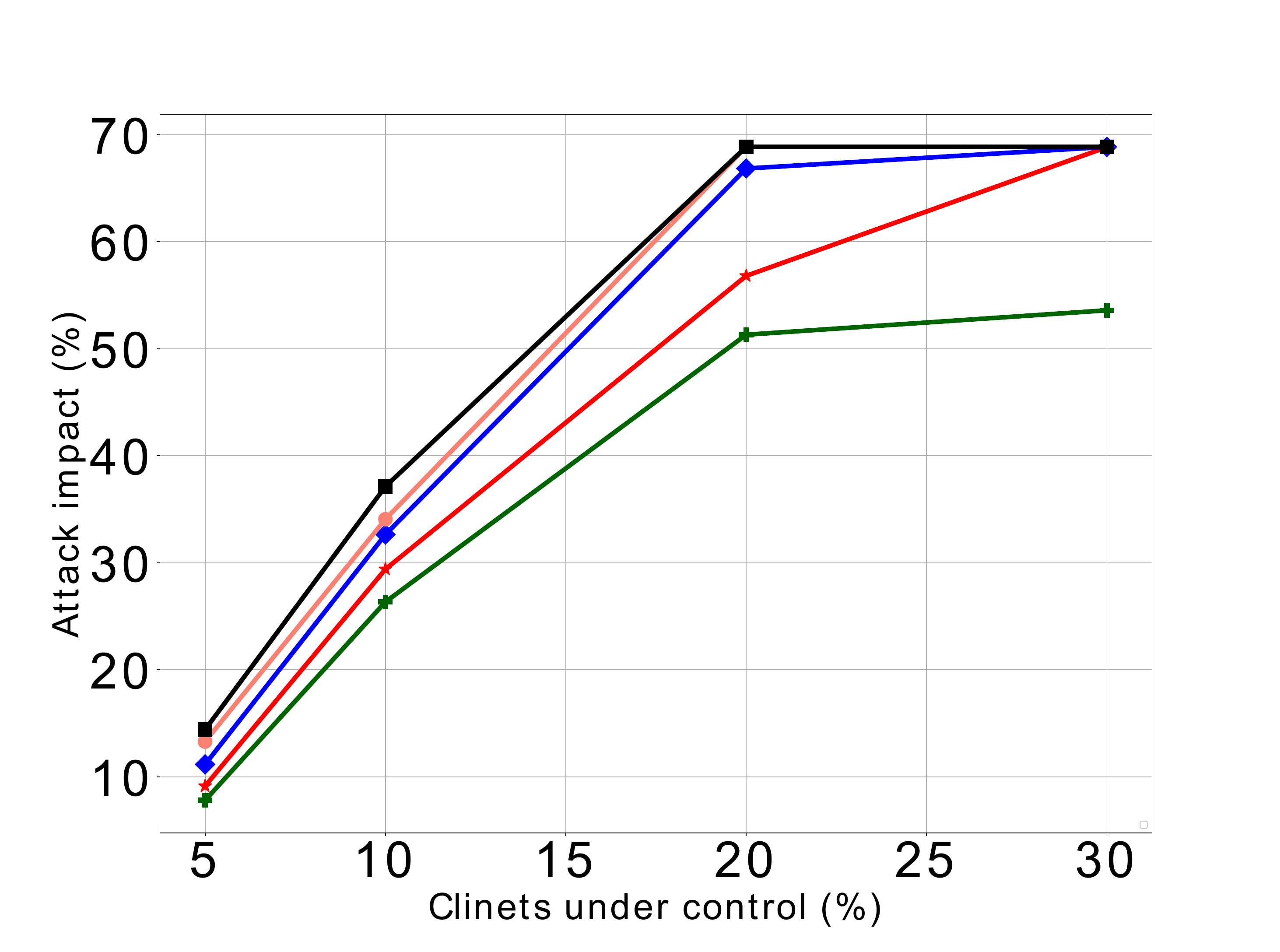}}
    \end{subfigmatrix} 
    \begin{subfigmatrix}{3}
    \centering
      \subfigure[FEMNIST + Median (No attack acc=84.29\%)]{
      \includegraphics[width=0.32\textwidth]{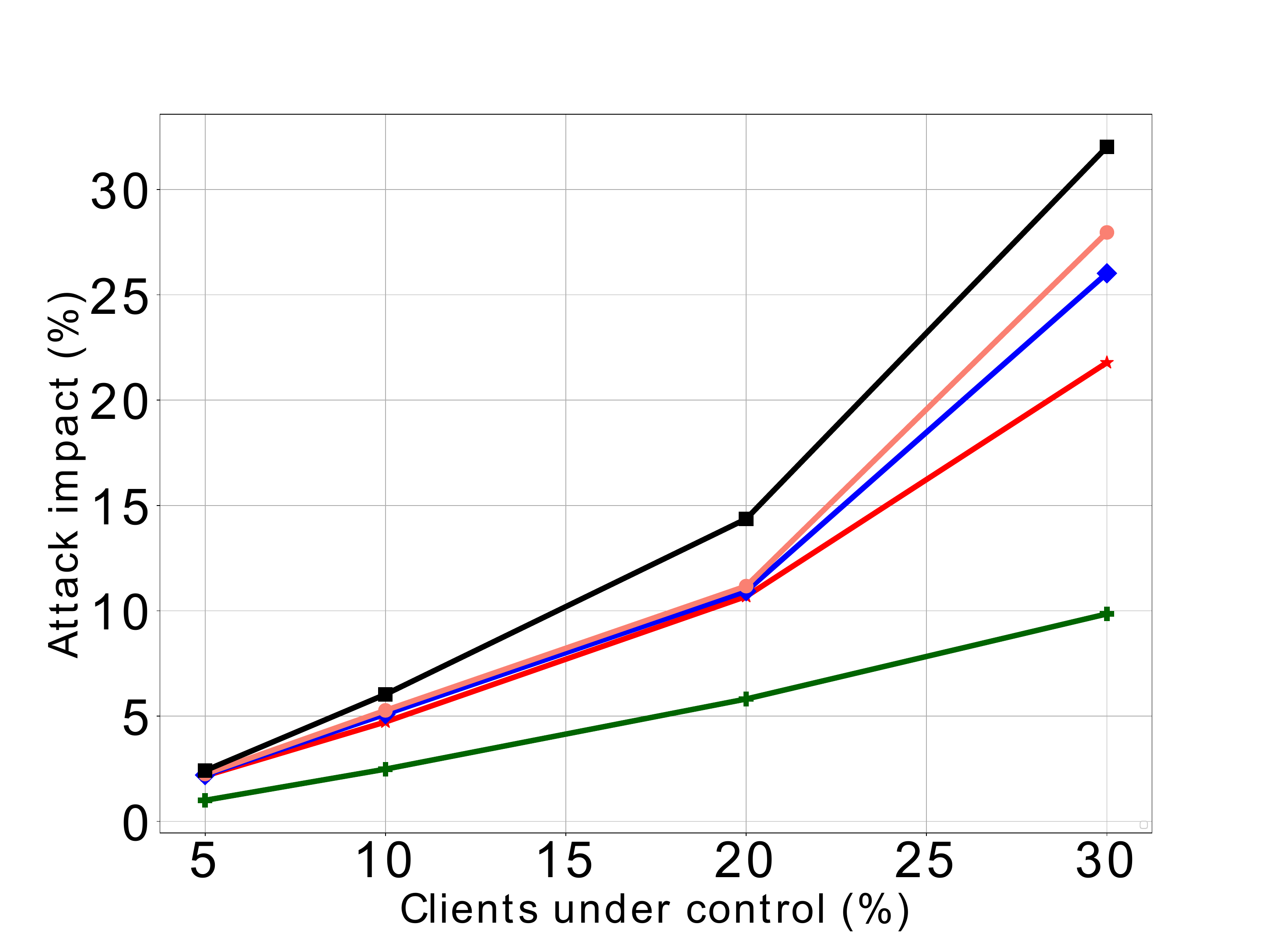}}
     \centering
      \subfigure[FEMNIST + NB $\tau=2.0$ (No attack acc=87.49\%)]{
      \includegraphics[width=0.32\textwidth]{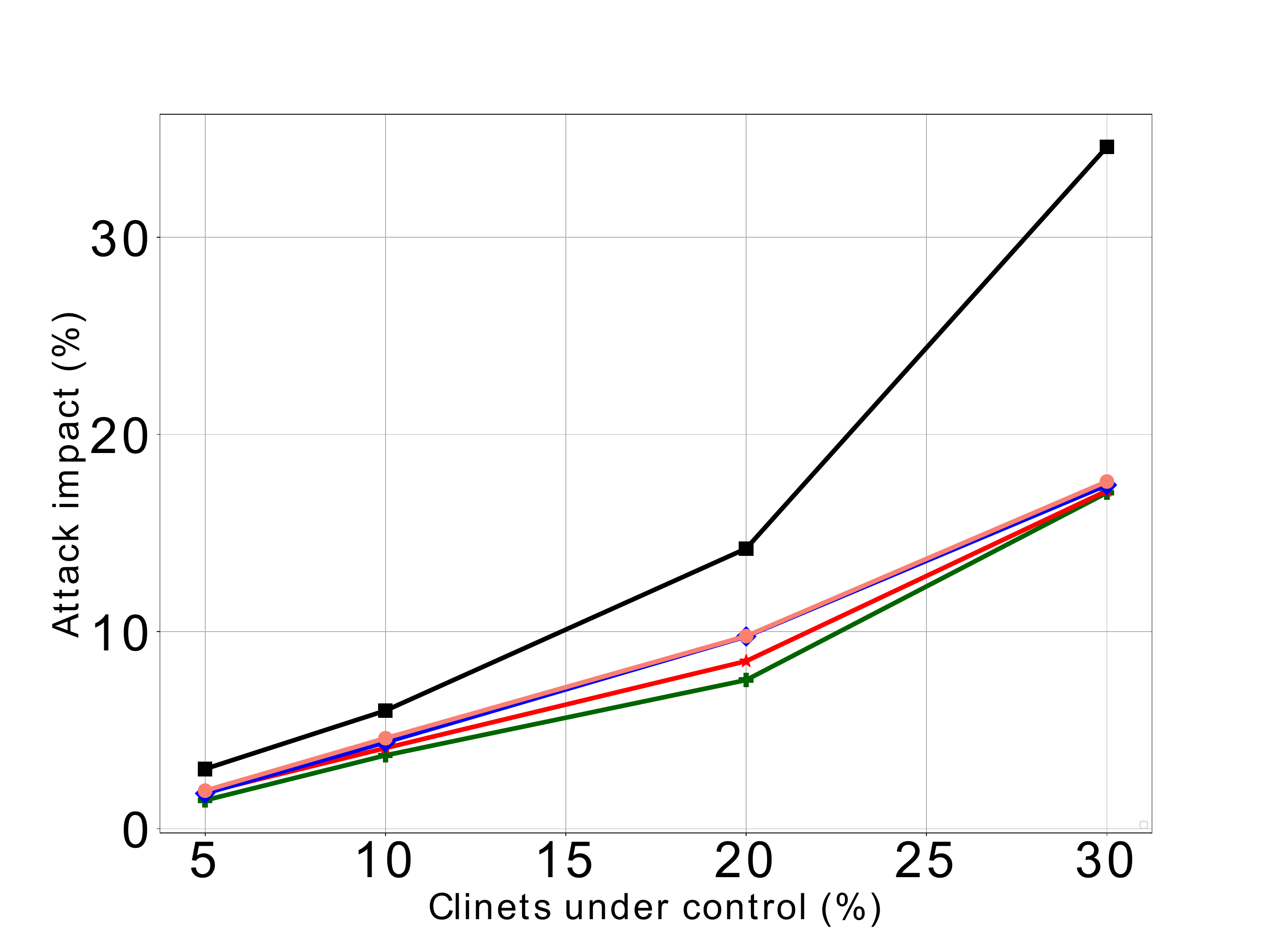}}
      \centering
      \subfigure[FEMNIST + NB $\tau=0.5$ (No attack acc=86.35\%)]{
      \includegraphics[width=0.32\textwidth]{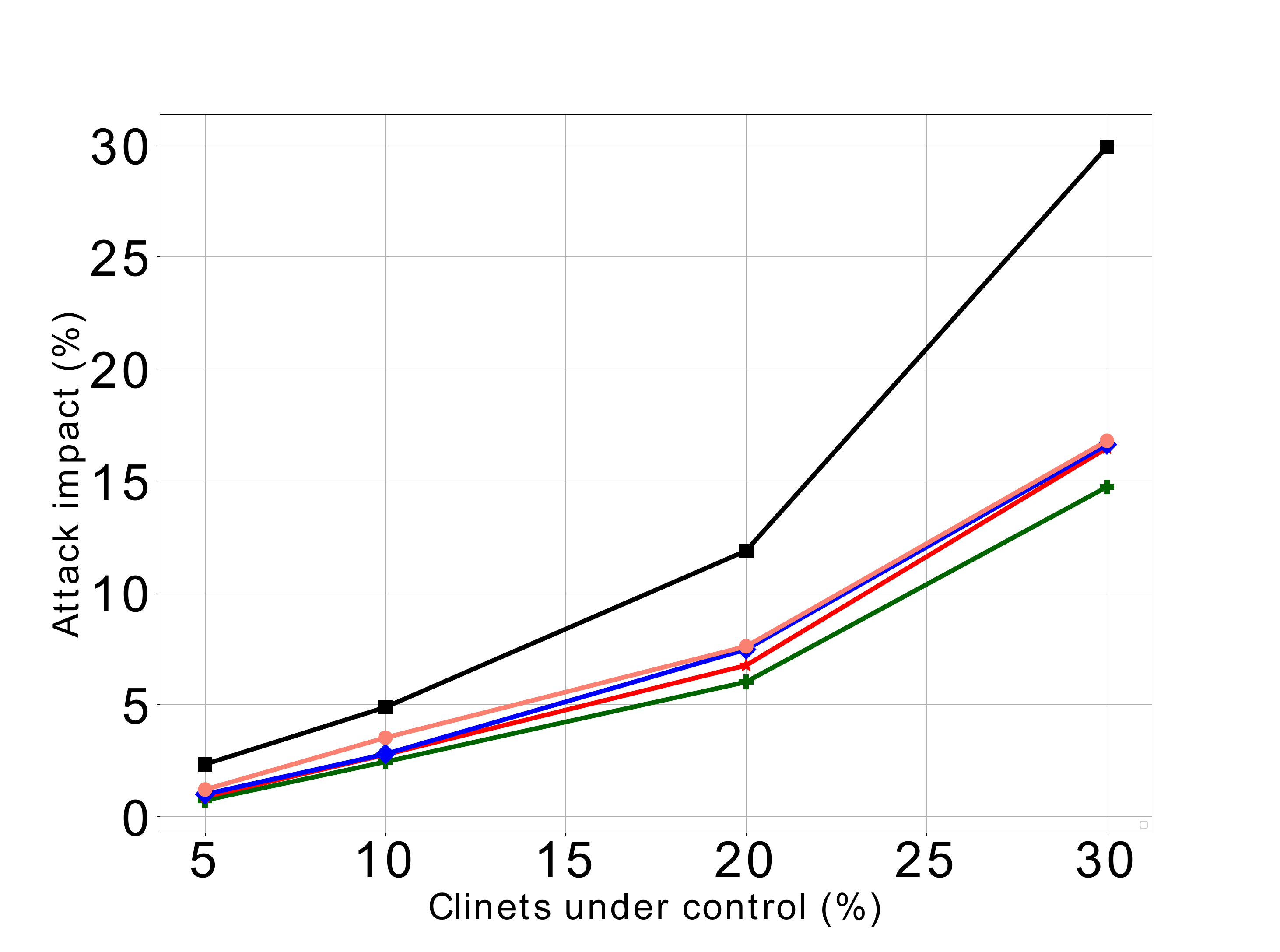}}
    \end{subfigmatrix} 
    \end{center}
    \caption{Attack impact ($I_{\theta}$) of the Norm-Bounding and Median aggregation rules in the presence of different adversaries. $\tau$ shows the $\ell_2$ threshold value that is used in Norm-Bounding aggreagtion. For hybrid attacks, we explore the impact of different numbers of compromised clients, specifically 0.5\% (5 clients), 0.3\% (3 clients), and 0.1\% (1 client) in CIFAR10 experiments and 0.5\% (17 clients), 0.3\% (11 clients), and 0.1\% (4 clients) in FEMNIST experiments.}
    \label{fig:agnostic_aware}
\end{figure*}

\subsection{Generating synthetic data using DDPM}
In Section~\ref{sec:hybridattack}, we explained the pipeline of our hybrid attack, which takes control of a few real clients and generates new synthetic data. In this section, we explain the details of this process for images of CIFAR10 and FEMNIST. To generate new samples, we use the following steps (similar to steps provided in Figure~\ref{fig:hybrid_attack}):

\paragraphb{Collecting the data of compromised clients.} We collect all the data samples of 0.1\%, 0.3\% and 0.5\% of first clients in both CIFAR10 and FEMNIST learning. For CIFAR10, we distribute the data in a non-iid fashion using Dirichlet distribution with parameter $\beta=0.5$. We saved the data assignment of the dataset and used this fixed distribution throughout our experiments. For CIFAR10, we collect the data samples of the first 1 (0.1\%), 3 (0.3\%), and 5 (0.5\%) of the clients. Figure~\ref{fig:Number_data_samples_CIFAR10_collection} shows the number of samples for each label (label 0 represents airplane images, label 1 represents car images, etc.) for our data collection. As we can see from this figure, when the attacker has only compromised 0.1\% of clients, it does not have access to any data samples of labels 3, 6, and 9. This means it cannot produce any new samples for these labels. For compromising 0.3\%, the adversary does not have access to any samples from label 9.  For FEMNIST, we also used the same generated data assignment (produce non-idd), and we collected the data samples of the first 4 (0.1\%), 7 (0.3\%), and 11 (0.5\%) of the clients.

\paragraphb{Generating new samples using DDPM}
We use the code provided in~\cite{ddpm_implementation} to generate new samples for the hybrid attacks. This code implemented the denoising diffusion probabilistic model (DDPM)~\cite{NEURIPS2020_4c5bcfec} in PyTorch.  It is a transcribed code from the official Tensorflow version~\cite{ddpm_implementation_tensorflow}. It uses denoising score matching to estimate the gradient of the data distribution, followed by Langevin sampling to sample from the true distribution.
After collecting the data samples of compromised clients, we ran the DDPM on these images for each label separately to generate new samples. To train the diffusion model, we used a batch size of 8, learning rate of 0.00008, and 250 sampling size. To generate samples for CIFAR10, we used 2000 diffusion steps, and for FEMNIST we used 1000 diffusion steps.

Figure~\ref{fig:DDPM_airplane} shows some DDPM-generated samples when the adversary has compromised 1 (0.1\%), 3 (0.3\%), and 5 (0.5\%) of the clients in learning of CIFAR10 distributed over 1000 clients. Figure~\ref{fig:Number_data_samples_CIFAR10_collection} shows the number of samples for each label. From this Figure, we can see the adversary has access to 1, 6, and 10 images of airplanes by compromising 0.1\%, 0.3\%, and 0.5\% of the clients, respectively. In Figure~\ref{fig:DDPM_airplane}(a), we can see that the DDPM model memorized the only image it has, and it just tried to add randomness to it because it has access to only one image of an airplane. Moreover, in Figure~\ref{fig:DDPM_airplane}(b) and Figure~\ref{fig:DDPM_airplane}(c), we can see that the model can generate better samples as it has access to more images from the true distribution. 
\

\paragraphb{Data assignment for the injected fake clients.} In all the hybrid attacks experiments, we first create a large dataset of all synthetic images from all the labels. We create this dataset by generating 5 samples per label multiplied by the number of injected fake clients. Then we distributed this dataset over the fake clients in a non-iid fashion using Dirichlet distribution with parameter $\beta=0.5$ for both CIFAR10 and FEMNIST experiments. 
Next, for launching the model poisoning attacks provided in Section~\ref{sec:comp_attacks}, the adversary chooses 25 random fake clients for its optimization and creates its malicious updates. This process happens in each FL round based on the global parameters $\theta^t$.

\section{Experiments} \label{sec:Exp}
 In this section, we conduct experiments to evaluate the performance of different Byzantine robust aggregation rules under different adversaries, using the FEMNIST~\cite{DBLP:journals/corr/abs-1812-01097, DBLP:conf/ijcnn/CohenATS17} and CIFAR10~\cite{krizhevsky2009learning} datasets. We consider a range of malicious client percentages, including 5\%, 10\%, 20\%, and 30\%, and report the maximum test accuracy and the impact of various attacks on the global model. For each attack, we also report attack cost, the number of benign, compromised, and injected fake clients present in the FL training process.

We consider five different attack scenarios, ranging from injecting fake clients with no knowledge of the true data distribution to a scenario where the adversary can compromise benign clients and use their data to craft malicious updates. Additionally, we propose and evaluate three types of hybrid attacks, where the adversary first compromises a small number of real clients and then uses their data to generate synthetic samples using a DDPM, followed by injecting fake clients with the new data samples. We explore the impact of different numbers of compromised clients in these hybrid attacks, specifically 0.5\% (5 clients), 0.3\% (3 clients), and 0.1\% (1 client) in CIFAR10 experiments and 0.5\% (17 clients), 0.3\% (11 clients), and 0.1\% (4 clients) in FEMNIST experiments. We rank the attacks in terms of their impact on the global model accuracy, to better illustrate the spectrum of attacks.

In our analysis, the 'malicious ratio' is determined by dividing the total number of fake and compromised clients by the overall client count, which includes fake, compromised, and benign clients. For instance, in a CIFAR10 setup distributed over 1000 clients where the adversary controls 10\%, if only 1 client is compromised (999 benign), the adversary can add 111 fake clients. This results in a malicious ratio of $\frac{(1+111)}{(1+111+999)}$, equating to 10\%. Conversely, if 5 clients are compromised (995 benign), maintaining the same malicious ratio would allow for 106 fake clients, calculated as $\frac{(5+106)}{(5+106+995)}$ = 10\%.

It is worth noting that we omit the results of the standard aggregation rule, FedAvg, as it is known to be vulnerable to even a single malicious client~\cite{blanchard2017machine} and can result in the global test accuracy approaching random guessing.

\subsection{Attacking agnostic robust AGRs}
In this section, we evaluate the performance of agnostic robust AGRs, specifically Median and Norm-Bounding, under different adversary models. We specifically focus on a spectrum of adversaries who control different percentages of malicious clients.

\paragraphb{Median AGR.} We present our experimental results using the Median aggregation rule in Figure~\ref{fig:agnostic_aware} (a) and (d) for CIFAR10 and FEMNIST experiments, respectively. Detailed results, including the attack cost, the number of benign, compromised, and injected fake clients corresponding to each attack, are provided in Table~\ref{tab:cifar10_median} and Table~\ref{tab:FEMNIST_median} (in Appendix~\ref{app:tables}).

Our findings indicate that the most potent adversary, who has compromised real clients, exerts the most significant influence on the global model. For instance, on the CIFAR10 dataset with the Median as the AGR, an attack by 10\% (20\%) malicious clients reduces the model's accuracy to 33.10\% (10.61\%). This implies that the attacker first compromised 100 (200) clients out of the total clients participating in FL and launched the attack described in Section~\ref{sec:comp_attacks} to craft its malicious update. The costs of these attacks would be \$10,000 and \$20,000, respectively, making them quite expensive.

 On the other hand, fake clients, who do not have any knowledge about the benign clients' data distribution, have the least impact on the global model. For example, on CIFAR10 with Median as the AGR, an attack launched by 10\% (20\%) of malicious clients reduces the accuracy of the global model to 49.04\% (32.78\%). To accomplish this, the adversary must inject 112 (251) fake clients into the FL training, which incurs costs of \$112 and \$251, respectively, considerably cheaper than compromising attacks.

Hybrid attacks, positioned in the middle of the spectrum, reveal that if the hybrid adversary has access to more data (more compromised clients), they can inflict more significant damage on the global model's accuracy. For instance, in the CIFAR10 dataset with the Median as the AGR, a hybrid attack involving 20\% malicious clients, where the adversary has compromised 1, 3, and 5 clients while generating new instances and injecting 249, 247, and 244 new fake clients, can reduce the FL model's accuracy to 13.29\%, 11.71\%, and 11.49\%, respectively. These attacks cost \$250, \$300, and \$500, respectively, which is very close to the cost of the fake attacks. Similar observations are made for the FEMNIST dataset as well.

\paragraphb{Norm-Bounding AGR.}\label{sec:exp_normb}
We report the experimental results  of our experiments when the server applies Norm-Bounding with a threshold $\tau$ as the aggregation rule in Figure~\ref{fig:agnostic_aware} (b), (c), (e), and (f) for CIFAR10 and FEMNIST datasets with two thresholds $\tau=0.5$ and $\tau=2.0$. Additionally, we provide information about the cost of attack, the number of fake and compromised clients in each attack in Table~\ref{tab:cifar10_normb} and Table~\ref{tab:FEMNIST_normb} in Appendix~\ref{app:tables}.
Our results show that the Norm-Bounding aggregation rule has similar impacts on the global model's accuracy as the Median AGR, when faced with different types of attacks.
For example, on CIFAR10 with $\tau=0.5$, when the adversary controls 10\% of clients, the fake adversary can inject 112 fake clients (with a cost of \$112) and reduce the accuracy to 52.52\%; the hybrid attack who compromised 1 client and injected 110 clients (with a cost of \$111) reduces the accuracy to 49.46\%; the hybrid attacker who compromised 3 clients and injected 108 fake clients (with a cost of \$300) reduces the accuracy to 46.22\%; the hybrid attacker who compromised 5 clients and injected 106 clients (with a cost of \$500) reduces the accuracy to 44.79\%; and at the end of the spectrum, a powerful adversary who compromised 100 clients (with a cost of \$10,000) can reduce the accuracy to 41.73\%.

\paragraphb{Larger upper bounds in Norm-Bounding results in more damage to the global model.} 
In our experiments, we consider two thresholds for Norm-Bounding $\tau=0.5$ and $\tau=2.0$. Our results show that for a larger threshold bound ($\tau$), the adversary has a larger space to craft its malicious updates and have a more significant impact on the FL global model. For instance, on FEMNIST, the compromising adversary with 30\% malicious ratio causes the accuracy dropped by 34.58\% when $\tau=2.0$ while the accuracy drop for the same setting and $\tau=0.5$ is about 29.92\%. 

 Therefore, with larger norm thresholds for the Norm-Bounding aggregation rule, the attackers have more impact on the global model. Alternatively, If the server wants to use a smaller threshold, then the model will result in lower accuracy when there is no malicious client. For instance, on CIFAR10, with no malicious clients, Norm-Bounding with threshold $\tau=0.5$ results in 78.86\% while $\tau=2.0$ results in 83.68\%; 10\% compromised clients will result in losing of 37.13\% and 73.68\% for $\tau=0.5$ and $\tau=2.0$ respectively. Therefore, there is a trade-off for choosing a proper threshold for bounding the local updates based on the assumption of the number of malicious clients in FL training.

\begin{figure}[hbt!]
    \centering
    \includegraphics[width=0.99\columnwidth]{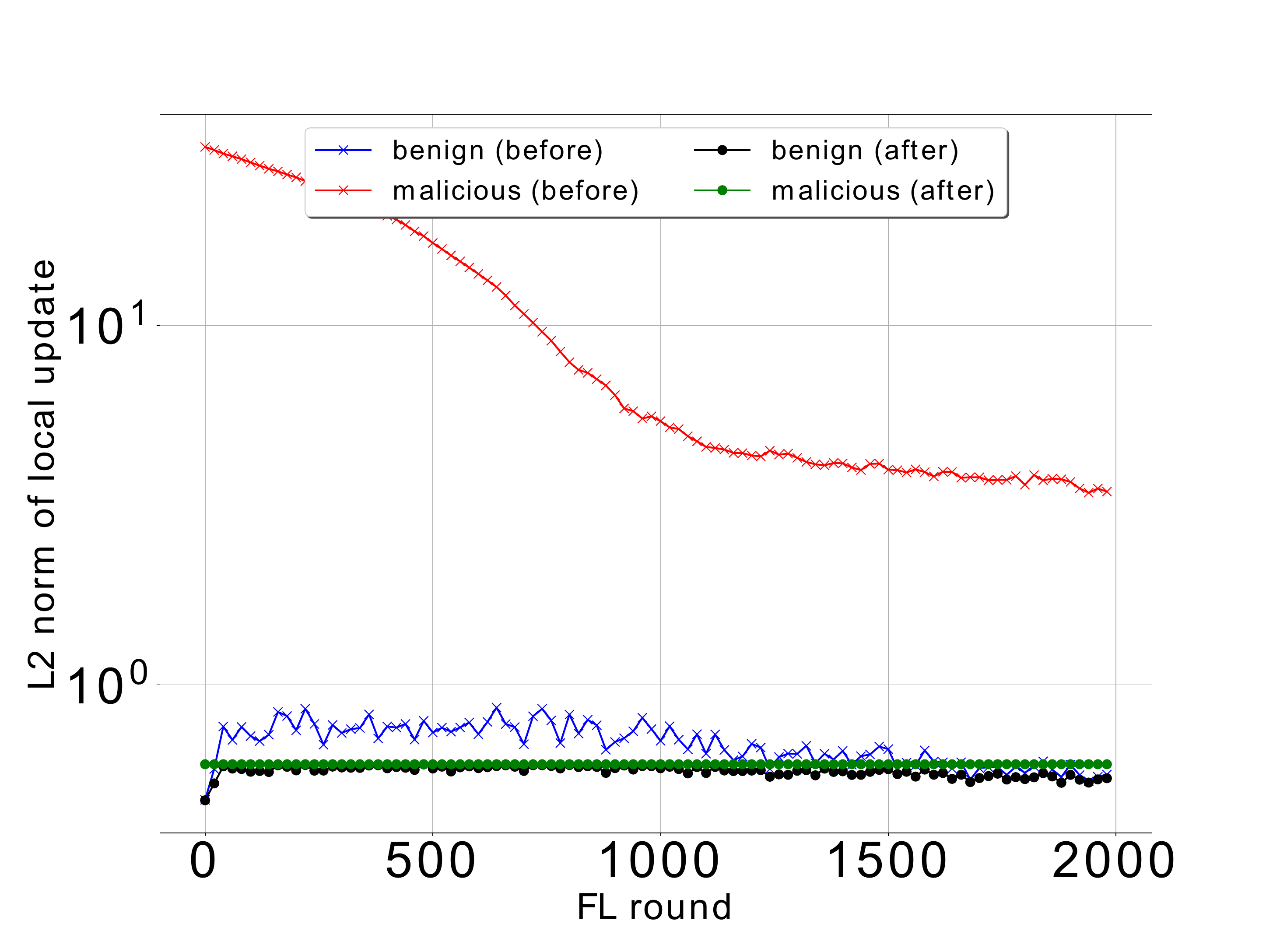}
    \caption{Local update norms throughout the FL training on CIFAR10 with 1000 benign clients and 112 fake clients (i.e., the adversary controls 10\% of total clients). In this figure, we can see that after FL round 1500, the malicious updates have a more considerable impact on the aggregation compared to benign updates because they have larger updates after norm bounding.}
    \label{fig:norm_b_CIFAR10_tau05}
\end{figure}

\paragraphb{Why can fake clients cause a significant attack impact for Norm-Bounding AGR?}
Figure~\ref{fig:norm_b_CIFAR10_tau05} shows the L2 norm of the updates (for malicious and benign updates for 10\% of malicious ratio in fake attack) before and after bounding the updates to $\tau=0.5$ for learning CIFAR10 throughout 2000 FL rounds. From this figure, we can see that when the global model starts to converge, the L2 norm of the local updates from benign updates becomes smaller than the threshold. For the updates that have norms smaller than the threshold, no change will be applied to them. However, on the other hand, the malicious updates are always greater than the threshold, so they are scaled down to have an L2 norm of $\tau$. In this figure, we can see that after FL round 1500, the malicious updates have a more considerable impact on the aggregation because they have larger updates.

\subsection{Attacking adaptive robust AGRs}
In this section, we conduct experiments to evaluate the robustness of adaptive Byzantine aggregation rules, specifically Trimmed-Mean~\cite{YinCRB18} and Multi-Krum~\cite{blanchard2017machine}, against a spectrum of adversaries who control varying percentages of malicious clients. In adaptive aggregation rules, we assume that the server has knowledge of the exact number of malicious clients in each FL round.

We report the performance of the Trimmed-Mean aggregation rule against different attacks in Table~\ref{tab:cifar10_trimmed} and Table~\ref{tab:FEMNIST_trimmed} (in Appendix~\ref{app:tables}) for FL models trained on the CIFAR10 and FEMNIST datasets, respectively, in the presence of 5\%, 10\%, 20\%, and 30\% of malicious clients. Similarly, Table~\ref{tab:cifar10_mkrum} and Table~\ref{tab:FEMNIST_mkrum} (in Appendix~\ref{app:tables}) show the attack impacts of different attacks when the server uses Multi-Krum as the aggregation rule for the CIFAR10 and FEMNIST datasets, respectively.

Our results indicate that adversaries who can compromise clients and use their data for attacks have the most significant impact on FL global models. For instance, on the CIFAR10 dataset, an adversary who has compromised 10\% (20\%) of clients, with a cost of \$10,000 (\$20,000), reduces the accuracy of FL by 55.83\% (73.66\%) and 49.29\% (60.37\%) with Trimmed-Mean and Multi-Krum, respectively. On the other hand, adversaries who can only inject fake clients into the FL training with no knowledge of the true data distribution have the lowest impact on global model accuracy. For instance, on the CIFAR10 dataset, an adversary who can inject 10\% (20\%) of clients, with a cost of \$112 (\$251), reduces the accuracy of FL by 39.78\% (51.17\%) and 1.32\% (3.55\%) with Trimmed-Mean and Multi-Krum, respectively.

Our experiments also show that the hybrid attack, which compromises only a few clients and use their data to produce more data samples for the fake clients, lies in the middle of the spectrum. The more clients are compromised, the more damage is done to the global accuracy. For instance, on the CIFARA10 training, a hybrid attacker who compromised 1 client, i.e., 0.1\% of total clients, and can inject 110 clients (in total 10\% malicious ratio) with a cost of \$111, can reduce the accuracy of the FL model by 54.24\% and 35.2\% for Trimmed-Mean and Multi-Krum respectively. While if the hybrid attacker compromised more clients (5 clients) and injected 106 clients (in total 10\% malicious ratio), with a cost of \$500, it can reduce the FL global accuracy by 55.77\% and 47.79\% for Trimmed-Mean and Multi-Krum, respectively.

Additionally, we also noticed that the Trimmed-Mean and Norm-Bounding (with $\tau=0.5$) are more vulnerable to injected fake clients with no knowledge about the true distribution of the training datasets. On the other hand, Multi-Krum can easily detect them and exclude them from aggregation. For instance, on CIFAR10, 10\% of injected fake clients (with \$112 attack cost) can reduce the accuracy of the model by 26.34\% and 39.78\% with Norm-Bounding and Trimmed-Mean as the aggregation rule, respectively. On the other hand, Multi-Krum only loses 1.32\% with the presence of this number of injected fake clients.

\subsection{Exploring Alternative Image Generation Baselines}
In our previous experiments, we exclusively utilized Denoising Diffusion Probabilistic Models (DDPM) for generating new images in our hybrid poisoning attacks. The choice of DDPM was motivated by its proven effectiveness in creating high-quality synthetic data. These models excel in producing data that closely resembles real datasets, a critical factor for successful poisoning attacks in federated learning. To broaden our investigation, we now consider alternative baselines for generating poisoning updates: Deep Convolutional Generative Adversarial Network (DCGAN) and the direct use of original images from compromised clients.

\begin{table*}[hbt!]
\caption{Global model accuracy and attack Impact ($I_{\theta}$) on CIFAR10 using Different Image Generation Methods in a hybrid attack (adversary has control of 10\% of the clients). We consider Median (No Attack Accuracy = 76.05\%), Trimmed-Mean (No Attack Accuracy = 83.66\%), Multi-Krum (No Attack Accuracy = 83.44\%), and Norm-Bounding ($\tau=0.5$) (No Attack Accuracy = 78.86\%) as AGRs.} \label{tab:GAN}
\centering
\footnotesize
\resizebox{\textwidth}{!}{
\begin{tabular} {|c|c|c||c|c|}
  \hline
     Attack Type & Generative Method &  AGR  & Accuracy (\%) & Attack Impact (\%) \\ \hline

     \multirow{12}{*}{\parbox{1.5cm}{Hybrid \\ comp: 0.5\%  \\ \#benign: 995 \\ \#comp: 5 \\ \#fake: 106}} & \multirow{4}{*}{No Generative Model} & Median & 65.09 ($\pm$ 2.796) & 10.96 ($\pm$ 2.976) \\ \cline{3-5}    
    &  & Trimmed-Mean & 74.04 ($\pm$ 2.861) & 9.62  ($\pm$ 2.861) \\ \cline{3-5}
   &  & Multi-Krum  & 76.92 ($\pm$ 1.650) & 6.52 ($\pm$ 1.650) \\ \cline{3-5}
    &  & Norm-Bounding ($\tau=0.5$) &  73.61 ($\pm$ 2.924) & 5.07 ($\pm$ 2.924)  \\ \cline{2-5}

     & \multirow{4}{*}{DCGAN} & Median  &  34.72 ($\pm$ 1.224)   &  41.33 ($\pm$ 1.224)  \\ \cline{3-5}    
    &  & Trimmed-Mean&  28.94 ($\pm$ 1.024) & 54.72 ($\pm$ 1.024) \\ \cline{3-5}
    &  & Multi-Krum  &  40.15 ($\pm$ 1.110) & 43.51 ($\pm$ 1.110) \\ \cline{3-5}
    &  & Norm-Bounding ($\tau=0.5$) &  49.41 ($\pm$ 2.528) &  29.45 ($\pm$ 2.528) \\
   \cline{2-5}

    & \multirow{4}{*}{DDPM} & Median & 33.48 ($\pm$ 1.337)  &  42.57 ($\pm$ 1.337) \\ \cline{3-5}    
     &  & Trimmed-Mean& 27.89 ($\pm$ 0.909) & 55.77 ($\pm$ 0.909) \\ \cline{3-5}
    &  & Multi-Krum & 35.65 ($\pm$ 0.956) & 47.79 ($\pm$ 0.956)  \\ \cline{3-5}
    &  & Norm-Bounding ($\tau=0.5$) & 44.79 ($\pm$ 2.152) & 34.07 ($\pm$ 2.152)   \\ \hline
\end{tabular}
}
\end{table*}

We adopted DCGAN, a well-regarded GAN variant, for its capability in generating high-fidelity images. Using the open-source DCGAN-PyTorch implementation~\cite{GAN_implementation}, we generated new samples for our hybrid attacks. DCGAN, noted for its effective network design, primarily comprises convolution layers, eschewing max pooling or fully connected layers. It utilizes strided and transposed convolutions for downsampling and upsampling, respectively. We paid particular attention to optimizing the network's convolutional features to ensure that the generated images were diverse and representative of the compromised data.

For training, we set the learning rate at 0.0002 and the slope of LeakyReLU at 0.2. We opted for the Adam optimizer (with $\beta1=0.5$ and $\beta2=0.999$) as suggested in~\cite{radford2015unsupervised}, and ran the model for 4000 epochs.

Figure~\ref{fig:DDPM_airplane}, second row, displays some DCGAN-generated samples under scenarios where 1 (0.1\%), 3 (0.3\%), and 5 (0.5\%) of clients in a CIFAR10 setting, distributed over 1000 clients, were compromised. A comparison between DDPM and DCGAN samples in Figure~\ref{fig:DDPM_airplane} reveals that DDPM more effectively captures the essence of training images and yields superior image quality.

As a direct comparison, we also considered a baseline scenario where the adversary employs original, unaltered images from compromised clients. This approach, lacking any generative modeling, serves as a control to assess the added value of complex image generation strategies in our hybrid attack framework.

Table~\ref{tab:GAN} presents our experimental results for CIFAR10, comparing the impact of Median, Multi-Krum, Trimmed-Mean, and Norm-Bounding (with $\tau=0.5$) aggregation rules. These experiments focus on a hybrid attack scenario where the adversary compromised 0.5\% (5 out of 1000) of the clients. In our experiments, we consider that the adversary controls 10\% of the FL clients including 5 compromised and 106 fake clients, as opposed to 995 benign clients.

From Table~\ref{tab:GAN}, it is evident that DCGAN is also highly effective in generating quality images that can be leveraged to poison the global model. The attack impacts using DCGAN are comparable to those using DDPM. For instance, under the Multi-Krum aggregation rule, the global model's accuracy is reduced by 43.51\% and 47.79\% using DCGAN and DDPM, respectively. This indicates that while both methods are potent, DDPM may have a slight edge in generating more realistic synthetic data, especially with limited and noisy datasets. The quality of generated data is pivotal in our hybrid attack model, as it significantly influences the attack's effectiveness.

The results also highlight that replicating data samples from compromised clients leads to a diminished attack impact. This is attributed to the high similarity in updates from these malicious clients. For instance, using replicated data under the Multi-Krum  only reduces the global model's accuracy by 6.52\%. This experiment demonstrates a higher variance in accuracy across all aggregation rules, further emphasizing the need for a generative model to achieve a higher and more stable attack impact.

A key element of our approach is the use of diverse reference updates, as delineated in step 4 of Figure~\ref{fig:hybrid_attack}. Diversity in these updates is critical for effectively crafting malicious updates that can circumvent various robust aggregation rules. For example, in the case of Multi-Krum, Trimmed-Mean, and Median rules, DYN-OPT algorithm generates poison updates based on equations~\ref{dyn_opt_mkrum} and~\ref{dyn_opt_trmean}. This diversity is essential in a realistic adversarial scenario, where compromised clients have varying data distributions, thereby enabling the most effective poisoned model update.

In scenarios where reference model updates are too similar, such as when replicating data points, the adversary's capability to explore and find impactful poisoned updates is substantially reduced. Uniformity in updates not only increases detection risk by the server but also diminishes the influence of the poisoned updates on the global model. This further substantiates the advantage of employing advanced generative models in conducting effective and stealthy hybrid attacks in federated learning environments.
\section{Conclusions}
In conclusion, this work presents a comprehensive study of the poisoning threats to FL by considering a spectrum of adversaries and robust AGRs. We discussed the importance of considering a spectrum of adversarial models, rather than focusing on the extremes, as it provides a more realistic understanding of the poisoning threat to various types of FL applications. We identify a hybrid adversary model where an adversary first compromises a few real clients and use their data to generate more data samples for the fake clients to mount a large-scale attack. For such a hybrid adversary, we propose a novel hybrid attack that leverages the denoising diffusion probabilistic model (DDPM) to generate new samples from a small number of compromised clients.



\bibliographystyle{IEEEtran}
\bibliography{ref}

\appendix
\section{Auxiliary results of model poisoning attacks against aware AGRs} \label{app:tables}
In this section, we present the results of our experiments for using different AGRs.
For each attack, we report the number of benign, compromised, and injected fake clients present in the FL training process.

\begin{table*}[hbt!]
\caption{Attack impact ($I_{\theta}$) and maximum test accuracy ($A_{\theta}^M$) of the Median for training on CIFAR10 distributed over 1000 initial clients in the presence of different adversaries. } \label{tab:cifar10_median}
\centering
\footnotesize
\resizebox{\textwidth}{!}{
\begin{tabular} {|c|c|c|c|c|c|c||c|c|}
  \hline
  AGR & Attack Type &  \parbox{1.1cm}{Malicious Rate} &  \parbox{1cm}{Number of Benign Clients} & \parbox{1cm}{Number of Compromised Clients}  & \parbox{1cm}{Number of \\Injected Fake Clients \\}  & \parbox{1.1cm}{Attack Cost (\$)} & Accuracy (\%) & Attack Impact (\%) \\ \hline 
  
  \multirow{20}{*}{\parbox{1cm}{Median (No attack acc = 76.05\%)}}  & 
  \multirow{4}{*}{Fake} & 5\% & 1000 & 0 & 53 & 53 & 63.94 ($\pm$ 1.253)  &12.11 ($\pm$ 1.253)  \circledE{5} \\ \cline{3-9}    
    &  & 10\% & 1000 & 0 & 112 & 112 & 49.04 ($\pm$ 0.649) &  27.01 ($\pm$ 0.649) \circledE{5}\\ \cline{3-9}
    &  & 20\% & 1000 & 0 & 251 & 251 & 32.78 ($\pm$ 0.699) & 43.27 ($\pm$ 0.699) \circledE{5} \\ \cline{3-9}
   &  & 30\% & 1000 & 0 & 429 & 429 & 25.41 ($\pm$ 4.937)  & 50.64 ($\pm$ 4.937) \circledE{5} \\ \cline{2-9} 

    & \multirow{4}{*}{\vtop{\hbox{\strut Hybrid} \hbox{\strut comp: 0.1\%}}} & 5\% & 999 & 1 & 52   & 100 & 49.08 ($\pm$ 1.131) & 26.97 ($\pm$ 1.131)  \circledD{4} \\ \cline{3-9} 
    
    &  & 10\% & 999 & 1 &  110 & 111 & 33.53 ($\pm$ 0.902) & 42.52 ($\pm$ 0.902) \circledC{3} \\ \cline{3-9}
    
   &  & 20\% & 999 & 1 &  249 &  250 & 13.29 ($\pm$ 6.026) & 62.76 ($\pm$ 6.026) \circledD{4} \\ \cline{3-9}
   
   &  & 30\% & 999 & 1 &  428 & 429 & 10.03 ($\pm$ 0.536) & 66.02 ($\pm$ 0.536) \circledD{4} \\ \cline{2-9} 

    & \multirow{4}{*}{\vtop{\hbox{\strut Hybrid} \hbox{\strut comp: 0.3\%}}} & 5\% & 997 & 3 & 50 & 300 & 48.85 ($\pm$ 1.258) &  27.20 ($\pm$ 1.258)  \circledC{3}\\ \cline{3-9}    
    &  & 10\% & 997 & 3 & 108  & 300 & 34.36 ($\pm$ 0.892)  & 41.69 ($\pm$ 0.892) \circledD{4} \\ \cline{3-9}
   &  & 20\% & 997 & 3 &  247 & 300 & 11.71 ($\pm$ 5.848)& 64.34 ($\pm$ 5.848) \circledC{3} \\ \cline{3-9}
   &  & 30\% & 997 & 3 &  425 & 428 & 10.00 ($\pm$ 0.000) & 66.05 ($\pm$ 0.000) \circledC{3}\\ \cline{2-9} 

    & \multirow{4}{*}{\vtop{\hbox{\strut Hybrid} \hbox{\strut comp: 0.5\%}}} & 5\% & 995 & 5 & 48 & 500 & 48.65 ($\pm$ 1.654) & 27.40 ($\pm$ 1.654)  \circledB{2}\\ \cline{3-9}    
    &  & 10\% & 995 & 5 &  106 & 500 & 33.48 ($\pm$ 1.337) &  42.57 ($\pm$ 1.337) \circledB{2} \\ \cline{3-9}
    &  & 20\% & 995 & 5 &  244 & 500 & 11.49 ($\pm$ 5.820) & 64.56 ($\pm$ 5.820) \circledB{2}  \\ \cline{3-9}
    &  & 30\% & 995 & 5 &  422 & 500 & 10.00 ($\pm$ 0.000)  & 66.05 ($\pm$ 0.000) \circledB{2} \\ \cline{2-9}

 & \multirow{4}{*}{Comp} & 5\% & 950 & 50 & 0 & 5,000 & 48.01 ($\pm$ 0.598) & 28.04 ($\pm$ 0.598) \circledA{1} \\ \cline{3-9} 
    &  & 10\% & 900 & 100 & 0 & 10,000 & 33.10 ($\pm$ 1.166) & 42.95 ($\pm$ 1.166) \circledA{1}\\ \cline{3-9}
    &  & 20\% & 800 & 200 & 0 & 20,000 & 10.61 ($\pm$ 1.669) & 65.44 ($\pm$ 1.669) \circledA{1} \\ \cline{3-9}
  &  & 30\% & 700 & 300 & 0 & 30,000 & 10.00 ($\pm$ 0.000) &  66.05 ($\pm$ 0.000) \circledA{1} \\ \hline
\end{tabular}}
\end{table*}
\begin{table*}[hbt!]
\caption{Attack impact ($I_{\theta}$) and maximum test accuracy ($A_{\theta}^M$) of the Median for training on FEMNIST distributed over 3400 initial clients in the presence of different adversaries.} \label{tab:FEMNIST_median}
\centering
\footnotesize
\resizebox{\textwidth}{!}{
\begin{tabular} {|c|c|c|c|c|c|c||c|c|}
  \hline
   AGR & Attack Type &  \parbox{1.1cm}{Malicious Rate} &  \parbox{1cm}{Number of Benign Clients} & \parbox{1cm}{Number of Compromised Clients}  & \parbox{1cm}{Number of \\Injected Fake Clients \\} & \parbox{1.1cm}{Attack Cost (\$)} & Accuracy (\%) & Attack Impact (\%) \\ \hline

  \multirow{20}{*}{\parbox{1cm}{Median (No attack acc = 84.29\%)}} & \multirow{4}{*}{Fake} &  5\% & 3400 & 0 &  179 & 179 & 83.29 ($\pm$ 0.146) &  1.00 ($\pm$ 0.146) \circledE{5} \\ \cline{3-9}    
     &  & 10\% & 3400 & 0 &  378 & 378 & 81.81 ($\pm$ 0.109)& 2.48 ($\pm$ 0.109) \circledE{5}\\ \cline{3-9}
    &  & 20\% & 3400 & 0 &  850 & 850 & 78.48 ($\pm$ 0.223)& 5.81 ($\pm$ 0.223) \circledE{5} \\ \cline{3-9}
     &  & 30\% & 3400 & 0 &  1458 & 1,458 & 74.44 ($\pm$ 0.574) & 9.85 ($\pm$ 0.574) \circledE{5}\\ \cline{2-9} 

& \multirow{4}{*}{\vtop{\hbox{\strut Hybrid} \hbox{\strut comp: 0.1\%}}} & 5\% & 3396 & 4 & 175 & 400 & 82.13 ($\pm$ 0.126) &  2.16 ($\pm$ 0.126) \circledD{4}\\ \cline{3-9}    
     &  & 10\% & 3396 & 4 & 374 & 400 & 79.57 ($\pm$ 0.275) & 4.72 ($\pm$ 0.275) \circledD{4}\\ \cline{3-9}
    &  & 20\% & 3396 & 4  & 845 &  849 & 73.61 ($\pm$ 0.756) & 10.68 ($\pm$ 0.756) \circledD{4} \\ \cline{3-9}
    &  & 30\% & 3396 & 4 & 1452 & 1,456 & 62.51 ($\pm$ 4.007) & 21.78 ($\pm$ 4.007) \circledD{4} \\ \cline{2-9} 

    & \multirow{4}{*}{\vtop{\hbox{\strut Hybrid} \hbox{\strut comp: 0.3\%}}} & 5\% & 3389 & 11 &  168 & 1,100 & 82.09 ($\pm$ 0.335)  & 2.20 ($\pm$ 0.335)   \circledC{3}\\ \cline{3-9}    
    &  & 10\% & 3389 & 11 & 366 & 1,100 & 79.20 ($\pm$ 0.194) & 5.09 ($\pm$ 0.194) \circledC{3}\\ \cline{3-9}
     &  & 20\% & 3389 & 11 & 837 & 1,100 & 73.36 ($\pm$ 0.989) & 10.93 ($\pm$ 0.989) \circledC{3} \\ \cline{3-9}
     &  & 30\% & 3389 & 11 & 1442 & 1,453 & 58.27 ($\pm$ 6.189) & 26.02 ($\pm$ 6.189) \circledC{3} \\ \cline{2-9} 

    & \multirow{4}{*}{\vtop{\hbox{\strut Hybrid} \hbox{\strut comp: 0.5\%}}} & 5\% & 3383 & 17 &  162 & 1,700 & 82.04 ($\pm$ 0.310)  &  2.25 ($\pm$ 0.310) \circledB{2} \\ \cline{3-9}    
     &  & 10\% & 3383 & 17 & 359 & 1,700 & 79.02 ($\pm$ 0.326) & 5.27 ($\pm$ 0.326)  \circledB{2}\\ \cline{3-9}
     &  & 20\% & 3383 & 17 & 829 & 1,700 & 73.12 ($\pm$ 0.333) & 11.17 ($\pm$ 0.333) \circledB{2} \\ \cline{3-9}
    &  & 30\% & 3383 & 17 & 1433 & 1,700 & 56.33 ($\pm$ 3.858) & 27.96 ($\pm$ 3.858) \circledB{2} \\ \cline{2-9}

     & \multirow{4}{*}{Comp} & 5\% & 3230 & 170 & 0 & 17,000 & 81.88 ($\pm$ 0.247)  &  2.41 ($\pm$ 0.247) \circledA{1}\\ \cline{3-9}    
     &  & 10\% & 3060 & 340 & 0 & 34,000 & 78.26 ($\pm$ 0.214) & 6.03 ($\pm$ 0.214) \circledA{1} \\ \cline{3-9}
    &  & 20\% & 2720 & 680 & 0 & 68,000 & 69.93 ($\pm$ 0.481) & 14.36 ($\pm$ 0.481) \circledA{1} \\ \cline{3-9}
    &  & 30\% & 2380 & 1020 & 0 & 102,000 & 52.27 ($\pm$ 1.458) & 32.02 ($\pm$ 1.458) \circledA{1} \\ 
   \hline 
\end{tabular}}
\end{table*}

\begin{table*}[hbt!]
\caption{Attack impact ($I_{\theta}$) and maximum test accuracy ($A_{\theta}^M$) of the Trimmed-Mean for training on CIFAR10 distributed over 1000 initial clients in the presence of different adversaries.} \label{tab:cifar10_trimmed}
\vspace{-5pt}
\centering
\footnotesize
\resizebox{\textwidth}{!}{
\begin{tabular} {|c|c|c|c|c|c|c||c|c|}
  \hline
     AGR & Attack Type &  \parbox{1.1cm}{Malicious Rate} &  \parbox{1cm}{Number of Benign Clients} & \parbox{1cm}{Number of Compromised Clients}  & \parbox{1cm}{Number of \\Injected Fake Clients \\}  & \parbox{1.1cm}{Attack Cost (\$)} & Accuracy (\%) & Attack Impact (\%) \\ \hline  
  
  \multirow{20}{*}{\parbox{1cm}{Trimmed-Mean (No attack acc = 83.66\%)}} &
  \multirow{4}{*}{Fake} & 5\% & 1000 & 0 & 53 & 53 & 59.95 ($\pm$ 0.617) &  23.71 ($\pm$ 0.617) \circledE{5}\\ \cline{3-9}    
     &  & 10\% & 1000 & 0 & 112 & 112 & 43.88 ($\pm$ 0.334)  & 39.78 ($\pm$ 0.334) \circledE{5} \\ \cline{3-9}
     &  & 20\% & 1000 & 0 & 251 & 251 & 32.49 ($\pm$ 0.451) & 51.17 ($\pm$ 0.451) \circledE{5} \\ \cline{3-9}
     &  & 30\% & 1000 & 0 & 429 & 429 & 25.56 ($\pm$ 0.238) & 58.10 ($\pm$ 0.238) \circledE{5} \\ \cline{2-9} 

     & \multirow{4}{*}{\vtop{\hbox{\strut Hybrid} \hbox{\strut comp: 0.1\%}}} & 5\% & 999 & 1 & 52 & 100 & 50.19 ($\pm$ 2.791) & 33.47 ($\pm$ 2.791) \circledD{4}\\ \cline{3-9}    
     &  & 10\% & 999 & 1 &  110 & 111 & 29.42 ($\pm$ 1.481) & 54.24 ($\pm$ 1.481) \circledD{4}\\ \cline{3-9}
     &  & 20\% & 999 & 1 &  249 & 250 & 20.61 ($\pm$ 5.277) & 63.05 ($\pm$ 5.277) \circledD{4} \\ \cline{3-9}
     &  & 30\% & 999 & 1 &  428 & 429 & 10.00 ($\pm$ 1.188) & 73.66 ($\pm$ 1.188) \circledD{4}  \\ \cline{2-9} 

     & \multirow{4}{*}{\vtop{\hbox{\strut Hybrid} \hbox{\strut comp: 0.3\%}}} & 5\% & 997 & 3 & 50 & 300 & 47.78 ($\pm$ 0.928) &  35.88 ($\pm$ 0.928) \circledC{3}\\ \cline{3-9}    
      &  & 10\% & 997 & 3 & 108  & 300 & 28.56 ($\pm$ 1.071) & 55.10 ($\pm$ 1.071) \circledC{3} \\ \cline{3-9}
  &  & 20\% & 997 & 3 &  247 & 300 & 20.50 ($\pm$ 5.415) & 63.16 ($\pm$ 5.415) \circledC{3} \\ \cline{3-9}
     &  & 30\% & 997 & 3 &  425 & 428 & 10.01 ($\pm$ 0.209) & 73.65 ($\pm$ 0.209) \circledC{3} \\ \cline{2-9} 

     & \multirow{4}{*}{\vtop{\hbox{\strut Hybrid} \hbox{\strut comp: 0.5\%}}} & 5\% & 995 & 5 & 48 & 500 & 41.90 ($\pm$ 3.438)  & 41.76 ($\pm$ 3.438) \circledA{1}\\ \cline{3-9}    
     &  & 10\% & 995 & 5 &  106 & 500 & 27.89 ($\pm$ 0.909) &  55.77 ($\pm$ 0.909) \circledB{2} \\ \cline{3-9}
     &  & 20\% & 995 & 5 &  244 & 500 & 20.31 ($\pm$ 5.151) & 63.35 ($\pm$ 5.151) \circledB{2} \\ \cline{3-9}
    &  & 30\% & 995 & 5 &  422 & 500 & 10.00 ($\pm$ 0.180) & 73.66 ($\pm$ 0.180) \circledB{2} \\ \cline{2-9}

    & \multirow{4}{*}{Comp} & 5\% & 950 & 50 & 0 & 5,000 & 44.25 ($\pm$ 1.195) & 39.41 ($\pm$ 1.195) \circledB{2}\\ \cline{3-9}    
    &  & 10\% & 900 & 100 & 0 & 10,000 & 27.33 ($\pm$ 0.346) & 55.83 ($\pm$ 0.346) \circledA{1}\\ \cline{3-9}
    &  & 20\% & 800 & 200 & 0 & 20,000 & 10.00 ($\pm$ 4.130) & 73.66 ($\pm$ 4.130) \circledA{1} \\ \cline{3-9}
    &  & 30\% & 700 & 300 & 0 & 30,000 & 10.00 ($\pm$ 0.000)) & 73.66 ($\pm$ 0.000) \circledA{1} \\    \hline 
\end{tabular}}
\end{table*}

\begin{table*}[hbt!]
\caption{Attack impact ($I_{\theta}$) and maximum test accuracy ($A_{\theta}^M$) of the Multi-Krum for training on CIFAR10 distributed over 1000 initial clients in the presence of different adversaries.} \label{tab:cifar10_mkrum}
\centering
\vspace{-5pt}
\footnotesize
\resizebox{\textwidth}{!}{
\begin{tabular} {|c|c|c|c|c|c|c||c|c|}
  \hline

   AGR & Attack Type &  \parbox{1.1cm}{Malicious Rate} &  \parbox{1cm}{Number of Benign Clients} & \parbox{1cm}{Number of Compromised Clients}  & \parbox{1cm}{Number of \\Injected Fake Clients \\}  & \parbox{1.1cm}{Attack Cost (\$)} & Accuracy (\%) & Attack Impact (\%) \\ \hline

      \multirow{20}{*}{ \parbox{1cm}{Multi-Krum (No attack acc = 83.44\%)}} &
     \multirow{4}{*}{Fake} & 5\% & 1000 & 0 & 53 & 53 & 82.70 ($\pm$ 0.291)& 0.74 ($\pm$ 0.291) \circledE{5} \\ \cline{3-9}    
      &  & 10\% & 1000 & 0 & 112 & 112 & 82.12 ($\pm$ 0.227) & 1.32 ($\pm$ 0.227) \circledE{5} \\ \cline{3-9}
      &  & 20\% & 1000 & 0 & 251 & 251 & 79.89 ($\pm$ 0.226) & 3.55 ($\pm$ 0.226) \circledE{5} \\ \cline{3-9}
      &  & 30\% & 1000 & 0 & 429 & 429 & 75.29 ($\pm$ 0.256)&  8.15 ($\pm$ 0.256) \circledE{5} \\ \cline{2-9} 

        & \multirow{4}{*}{\vtop{\hbox{\strut Hybrid} \hbox{\strut comp: 0.1\%}}} & 5\% & 999 & 1 & 52 & 100 & 70.12 ($\pm$ 0.895)  & 13.32 ($\pm$ 0.895) \circledD{4}\\ \cline{3-9}    
     &  & 10\% & 999 & 1 &  110 & 111 & 48.24 ($\pm$ 2.371) & 35.20 ($\pm$ 2.371) \circledD{4} \\ \cline{3-9}
     &  & 20\% & 999 & 1 &  249 & 250 & 24.71 ($\pm$ 0.257) & 58.73 ($\pm$ 0.257) \circledD{4}\\ \cline{3-9}
     &  & 30\% & 999 & 1 &  428 & 429  & 20.22 ($\pm$ 0.539) & 63.22 ($\pm$ 0.539) \circledD{4} \\ \cline{2-9}

    & \multirow{4}{*}{\vtop{\hbox{\strut Hybrid} \hbox{\strut comp: 0.3\%}}} & 5\% & 997 & 3 & 50 & 300 & 62.65 ($\pm$ 0.725)&  20.79 ($\pm$ 0.725) \circledC{3} \\ \cline{3-9}    
     &  & 10\% & 997 & 3 & 108  & 300 & 36.70 ($\pm$ 2.188) & 46.74 ($\pm$ 2.188) \circledC{3} \\ \cline{3-9}
     &  & 20\% & 997 & 3 &  247 & 300 & 23.79 ($\pm$ 1.788) & 59.65 ($\pm$ 1.788) \circledC{3}  \\ \cline{3-9}
    &  & 30\% & 997 & 3 &  425 & 428 & 19.90 ($\pm$ 2.234) & 63.54 ($\pm$ 2.234) \circledC{3}\\ \cline{2-9}

    & \multirow{4}{*}{\vtop{\hbox{\strut Hybrid} \hbox{\strut comp: 0.5\%}}} & 5\% & 995 & 5 & 48 & 500 & 62.47 ($\pm$ 0.914)& 20.97 ($\pm$ 0.914) \circledB{2} \\ \cline{3-9}    
    &  & 10\% & 995 & 5 &  106 & 500 & 35.65 ($\pm$ 0.956)&  47.79 ($\pm$ 0.956) \circledB{2}   \\ \cline{3-9}
    &  & 20\% & 995 & 5 &  244 & 500 & 23.10 ($\pm$ 1.433)& 60.34 ($\pm$ 1.433)  \circledB{2}  \\ \cline{3-9}
    &  & 30\% & 995 & 5 &  422 & 500 & 19.86 ($\pm$ 0.619) & 63.58 ($\pm$ 0.619) \circledB{2} \\ \cline{2-9}

     & \multirow{4}{*}{Comp} & 5\% & 950 & 50 & 0 & 5,000 & 62.04 ($\pm$ 1.307) & 21.40 ($\pm$ 1.307) \circledA{1}  \\ \cline{3-9}    
     &  & 10\% & 900 & 100 & 0 & 10,000 & 34.15 ($\pm$ 0.660) & 49.29 ($\pm$ 0.660) \circledA{1} \\ \cline{3-9}
     &  & 20\% & 800 & 200 & 0 & 20,000 & 23.07 ($\pm$ 0.528) & 60.37 ($\pm$ 0.528) \circledA{1}  \\ \cline{3-9}
    &  & 30\% & 700 & 300 & 0 & 30,000 & 19.31 ($\pm$ 0.786) & 64.13 ($\pm$ 0.786) \circledA{1}  \\ 
   \hline 
\end{tabular}
}
\end{table*}

\begin{table*}[hbt!]
\caption{Attack impact ($I_{\theta}$) and maximum test accuracy ($A_{\theta}^M$) of the Trimmed-Mean for training on FEMNIST distributed over 3400 initial clients in the presence of different adversaries.} \label{tab:FEMNIST_trimmed}
\vspace{-5pt}
\centering
\footnotesize
\resizebox{\textwidth}{!}{
\begin{tabular} {|c|c|c|c|c|c|c||c|c|}
  \hline
    AGR & Attack Type &  \parbox{1.1cm}{Malicious Rate} &  \parbox{1cm}{Number of Benign Clients} & \parbox{1cm}{Number of Compromised Clients}  & \parbox{1cm}{Number of \\Injected Fake Clients \\}  & \parbox{1.1cm}{Attack Cost (\$)} & Accuracy (\%) & Attack Impact (\%) \\ \hline

  \multirow{20}{*}{\parbox{1cm}{Trimmed-Mean (No attack acc = 87.52\%)}}& \multirow{4}{*}{Fake} & 5\% &  3400 & 0 &  179 & 179 & 84.90 ($\pm$ 0.108) &  2.62 ($\pm$ 0.108) \circledE{5} \\ \cline{3-9}    
   &  & 10\% & 3400 & 0 &  378 & 378 &  82.64 ($\pm$ 0.135)& 4.88 ($\pm$ 0.135) \circledE{5} \\ \cline{3-9}
   &  & 20\% & 3400 & 0 &  850 & 850 &  78.04 ($\pm$ 0.198) & 9.48 ($\pm$ 0.198) \circledE{5} \\ \cline{3-9}
    &  & 30\% & 3400 & 0 &  1458 & 1,458 & 73.11 ($\pm$ 0.384) & 14.41 ($\pm$ 0.384) \circledE{5} \\ \cline{2-8} 

& \multirow{4}{*}{\vtop{\hbox{\strut Hybrid} \hbox{\strut comp: 0.1\%}}} & 5\% & 3396 & 4 & 175 & 400 & 84.04 ($\pm$ 0.223)  &  3.48 ($\pm$ 0.223) \circledD{4} \\ \cline{3-9}    
    &  & 10\% & 3396 & 4 & 374 & 400 & 80.44 ($\pm$ 0.672) & 7.08 ($\pm$ 0.672) \circledD{4} \\ \cline{3-9}
   &  & 20\% & 3396 & 4  & 845 & 849 & 72.09 ($\pm$ 1.114) & 15.43 ($\pm$ 1.114) \circledD{4}  \\ \cline{3-9}
   &  & 30\% & 3396 & 4 & 1452 & 1,456 & 58.28 ($\pm$ 0.699) & 29.24 ($\pm$ 0.699) \circledD{4} \\ \cline{2-9}

     & \multirow{4}{*}{\vtop{\hbox{\strut Hybrid} \hbox{\strut comp: 0.3\%}}} & 5\% & 3389 & 11 &  168 & 1,100 & 83.95 ($\pm$ 0.151) &  3.57 ($\pm$ 0.151) \circledC{3} \\ \cline{3-9}    
     &  & 10\% & 3389 & 11 & 366 & 1,100 & 79.38 ($\pm$ 0.313) &  8.14 ($\pm$ 0.313) \circledB{2}\\ \cline{3-9}
     &  & 20\% & 3389 & 11 & 837 & 1,100 & 70.48 ($\pm$ 0.815) &  17.07 ($\pm$ 0.815) \circledC{3} \\ \cline{3-9}
    &  & 30\% & 3389 &  11 & 1442 & 1,453 & 57.15 ($\pm$ 1.953) & 30.37 ($\pm$ 1.953) \circledC{3} \\ \cline{2-9}

    & \multirow{4}{*}{\vtop{\hbox{\strut Hybrid} \hbox{\strut comp: 0.5\%}}} & 5\% & 3383 & 17 &  162 & 1,700 & 83.73 ($\pm$ 0.248) & 3.79 ($\pm$ 0.248) \circledB{2}  \\ \cline{3-9}    
     &  & 10\% & 3383 & 17 & 359 & 1,700 & 79.75 ($\pm$ 0.659) & 7.77 ($\pm$ 0.659) \circledC{3} \\ \cline{3-9}
    &  & 20\% & 3383 &17 & 829 & 1,700 & 70.33 ($\pm$ 2.009) & 17.19 ($\pm$ 2.009) \circledB{2} \\ \cline{3-9}
     &  & 30\% & 3383 & 17 & 1433 & 1,700 & 54.20 ($\pm$ 2.420) & 33.32 ($\pm$ 2.420) \circledB{2} \\ \cline{2-9}

     & \multirow{4}{*}{Comp} & 5\% & 3230 & 170 & 0 & 17,000 & 83.51 ($\pm$ 0.183) & 4.01 ($\pm$ 0.183) \circledA{1}  \\ \cline{3-9}    
    &  & 10\% & 3060 & 340 & 0 & 34,000 & 78.71 ($\pm$ 0.498) & 8.81 ($\pm$ 0.498) \circledA{1} \\ \cline{3-9}
    &  & 20\% & 2720 & 680 & 0 & 68,000 & 68.13 ($\pm$ 2.040) & 19.39 ($\pm$ 2.040) \circledA{1} \\ \cline{3-9}
     &  & 30\% & 2380 & 1020 & 0 & 102,000 & 40.35 ($\pm$ 2.275) & 47.17 ($\pm$ 2.275) \circledA{1} \\ \hline
\end{tabular}
}
\end{table*}

\begin{table*}[hbt!]
\caption{Attack impact ($I_{\theta}$) and maximum test accuracy ($A_{\theta}^M$) of the Multi-Krum for training on FEMNIST distributed over 3400 initial clients in the presence of different adversaries.} \label{tab:FEMNIST_mkrum}
\vspace{-5pt}
\centering
\footnotesize
\resizebox{\textwidth}{!}{
\begin{tabular} {|c|c|c|c|c|c|c||c|c|}
  \hline
     AGR & Attack Type &  \parbox{1.1cm}{Malicious Rate} &  \parbox{1cm}{Number of Benign Clients} & \parbox{1cm}{Number of Compromised Clients}  & \parbox{1cm}{Number of \\Injected Fake Clients \\}  & \parbox{1.1cm}{Attack Cost (\$)} & Accuracy (\%) & Attack Impact (\%) \\ \hline

     \multirow{20}{*}{\parbox{1cm}{Multi-Krum (No attack acc = 87.45\%)}} & \multirow{4}{*}{Fake} & 5\% &  3400 & 0 &  179 & 179 & 87.25 ($\pm$ 0.064) &  0.20 ($\pm$ 0.064)  \circledE{5} \\ \cline{3-9}    
    &  & 10\% & 3400 & 0 &  378 & 378 & 87.11 ($\pm$ 0.066) & 0.34 ($\pm$ 0.066) \circledE{5} \\ \cline{3-9}
   &  & 20\% & 3400 & 0 &  850 & 850 & 86.58 ($\pm$ 0.178) & 0.87 ($\pm$ 0.178) \circledE{5} \\ \cline{3-9}
    &  & 30\% & 3400 & 0 &  1458 & 1,458 & 85.60 ($\pm$ 0.174) & 1.85 ($\pm$ 0.174) \circledE{5} \\ \cline{2-9} 

 & \multirow{4}{*}{\vtop{\hbox{\strut Hybrid} \hbox{\strut comp: 0.1\%}}   } & 5\% & 3396 & 4 & 175 & 400 & 86.02 ($\pm$ 0.176) &  1.43 ($\pm$ 0.176) \circledC{3}\\ \cline{3-9}    
   &  & 10\% & 3396 & 4 & 374 & 400 & 82.82 ($\pm$ 0.352) & 4.63 ($\pm$ 0.352) \circledD{4} \\ \cline{3-9}
    &  & 20\% & 3396 & 4  & 845 & 849 & 75.88 ($\pm$ 0.635) & 11.57 ($\pm$ 0.635) \circledD{4} \\ \cline{3-9}
    &  & 30\% & 3396 & 4 & 1452 & 1,456 & 62.23 ($\pm$ 1.825) & 25.22 ($\pm$ 1.825) \circledC{3} \\ \cline{2-9}

    & \multirow{4}{*}{\vtop{\hbox{\strut Hybrid} \hbox{\strut comp: 0.3\%}}} & 5\% & 3389 & 11 &  168 & 1,100 & 86.26 ($\pm$ 0.106) &  1.19 ($\pm$ 0.106) \circledD{4}\\ \cline{3-9}    
   &  & 10\% & 3389 & 11 & 366 & 1,100 & 81.58 ($\pm$ 0.223) & 5.87 ($\pm$ 0.223) \circledA{1} \\ \cline{3-9}
    &  & 20\% & 3389 & 11 & 837 & 1,100 & 73.97 ($\pm$ 0.582) & 13.48 ($\pm$ 0.582) \circledC{3} \\ \cline{3-9}
    &  & 30\% & 3389 & 11 & 1442 & 1,453 & 62.35 ($\pm$ 0.859) & 25.10 ($\pm$ 0.859) \circledD{4}  \\ \cline{2-9}  

    & \multirow{4}{*}{\vtop{\hbox{\strut Hybrid} \hbox{\strut comp: 0.5\%}}} & 5\% & 3383 & 17 &  162 & 1,700 & 85.87 ($\pm$ 0.126)  &  1.98 ($\pm$ 0.126) \circledB{2}\\ \cline{3-9}    
     &  & 10\% & 3383 &17 & 359 & 1,700 & 82.03 ($\pm$ 0.376) & 5.42 ($\pm$ 0.376) \circledC{3}\\ \cline{3-9}
    &  & 20\% & 3383 &17 & 829 & 1,700 & 71.71 ($\pm$ 2.148) & 15.74 ($\pm$ 2.148) \circledB{2} \\ \cline{3-9}
    &  & 30\% & 3383 & 17 & 1433 & 1,700 & 61.94 ($\pm$ 1.990) & 25.51 ($\pm$ 1.990) \circledB{2}  \\ \cline{2-9}

     & \multirow{4}{*}{Comp} & 5\% & 3230 & 170 & 0 & 17,000 &  85.46 ($\pm$ 0.113)   & 1.99 ($\pm$ 0.113) \circledA{1} \\ \cline{3-9}    
    &  & 10\% & 3060 &  340 & 0 & 34,000 & 81.73 ($\pm$ 0.390) & 5.72 ($\pm$ 0.390) \circledB{2}\\ \cline{3-9}
    &  & 20\% & 2720 & 680 & 0 & 68,000 & 69.39 ($\pm$ 1.597) & 18.06 ($\pm$ 1.597) \circledA{1}\\ \cline{3-9}
    &  & 30\% & 2380 & 1020 & 0 & 102,000 &  47.83 ($\pm$ 10.627) & 39.62 ($\pm$ 10.627) \circledA{1} \\
   \hline 
\end{tabular}
}
\end{table*}

\begin{table*}[hbt!]
\caption{Attack impact ($I_{\theta}$) and maximum test accuracy ($A_{\theta}^M$) of the Norm-Bounding for training on CIFAR10 distributed over 1000 initial clients in the presence of different adversaries.} \label{tab:cifar10_normb}
\centering
\footnotesize
\resizebox{\textwidth}{!}{
\begin{tabular} {|c|c|c|c|c|c|c||c|c|}
  \hline
  AGR & Attack Type &  \parbox{1.1cm}{Malicious Rate} &  \parbox{1cm}{Number of Benign Clients} & \parbox{1cm}{Number of Compromised Clients}  & \parbox{1cm}{Number of \\Injected Fake Clients \\}  & \parbox{1.1cm}{Attack Cost (\$)} & Accuracy (\%) & Attack Impact (\%) \\ \hline 
 \multirow{20}{*}{  \parbox{1cm}{Norm Bounding ($\tau=2.0$) (No attack acc = 83.68\%)}} & 
     \multirow{4}{*}{Fake} & 5\% & 1000 & 0 & 53 & 53 & 39.55 ($\pm$ 0.577)  & 44.13 ($\pm$ 0.577) \circledE{5} \\ \cline{3-9}    
     &  & 10\% & 1000 & 0 & 112 & 112 & 25.65 ($\pm$ 0.450) & 58.03 ($\pm$ 0.450) \circledE{5} \\ \cline{3-9}
    &  & 20\% & 1000 & 0 & 251 & 251 & 15.48 ($\pm$ 0.416) &  68.20 ($\pm$ 0.416) \circledE{5} \\ \cline{3-9}
    &  & 30\% & 1000 & 0 & 429 & 429 & 14.63 ($\pm$ 1.012) & 69.05 ($\pm$ 1.012) \circledE{5} \\ \cline{2-9}

    & \multirow{4}{*}{\vtop{\hbox{\strut Hybrid} \hbox{\strut comp: 0.1\%}}} & 5\% & 999 &  1 & 52 & 100 & 22.14 ($\pm$ 2.023) & 61.54 ($\pm$ 2.023) \circledD{4} \\ \cline{3-9}  
  &  & 10\% & 999 & 1 &  110 & 111 & 10.00 ($\pm$ 0.000) & 73.68 ($\pm$ 0.000) \circledD{4} \\ \cline{3-9}
     &  & 20\% & 999 & 1 &  249 & 250 & 10.00 ($\pm$ 0.000) &  73.68 ($\pm$ 0.000) \circledD{4} \\ \cline{3-9}
    &  & 30\% & 999 & 1 &  428 &  429 & 10.00 ($\pm$ 0.000) & 73.68 ($\pm$ 0.000) \circledD{4} \\ \cline{2-9} 

    & \multirow{4}{*}{\vtop{\hbox{\strut Hybrid} \hbox{\strut comp: 0.3\%}}} & 5\% & 997 & 3 & 50 & 300 & 12.28 ($\pm$ 2.067) & 71.40 ($\pm$ 2.067) \circledC{3} \\ \cline{3-9}    
    &  & 10\% & 997 & 3 & 108  & 300 & 10.00 ($\pm$ 0.000) & 73.68 ($\pm$ 0.000) \circledC{3}   \\ \cline{3-9}
    &  & 20\% & 997 & 3 &  247 & 300 & 10.00 ($\pm$ 0.000) & 73.68 ($\pm$ 0.000) \circledC{3} \\ \cline{3-9}
    &  & 30\% & 997 & 3 &  425 & 428 & 10.00 ($\pm$ 0.000) & 73.68 ($\pm$ 0.000) \circledC{3}   \\ \cline{2-9} 

    & \multirow{4}{*}{\vtop{\hbox{\strut Hybrid} \hbox{\strut comp: 0.5\%}} } & 5\% & 995 & 5 & 48 & 500 & 10.00 ($\pm$ 1.125) & 73.68 ($\pm$ 1.125) \circledB{2} \\ \cline{3-9}    
     &  & 10\% & 995 & 5 &  106 & 500 & 10.00 ($\pm$ 0.000) & 73.68 ($\pm$ 0.000) \circledB{2}  \\ \cline{3-9}
    &  & 20\% & 995 & 5 &  244 & 500 & 10.00 ($\pm$ 0.000) & 73.68 ($\pm$ 0.000) \circledB{2} \\ \cline{3-9}
    &  & 30\% & 995 & 5 &  422 & 500 & 10.00 ($\pm$ 0.000) & 73.68 ($\pm$ 0.000) \circledB{2}  \\ \cline{2-9}

     & \multirow{4}{*}{Comp} & 5\% & 950 &  50 & 0 & 5,000 &  10.00 ($\pm$ 1.046) & 73.68 ($\pm$ 1.046) \circledA{1}\\ \cline{3-9}    
    &  & 10\% & 900 & 100 & 0 & 10,000 & 10.00 ($\pm$ 0.000) & 73.68 ($\pm$ 0.000) \circledA{1}  \\ \cline{3-9}
    &  & 20\% & 800 & 200 & 0 & 20,000 & 10.00 ($\pm$ 0.000) & 73.68 ($\pm$ 0.000) \circledA{1} \\ \cline{3-9}
    &  & 30\% & 700 & 300 & 0 & 30,000 & 10.00 ($\pm$ 0.000) & 73.68 ($\pm$ 0.000) \circledA{1} \\ \cline{1-9}

     \multirow{20}{*}{ \parbox{1cm}{Norm Bounding ($\tau=0.5$) (No attack acc = 78.86\%)} } & 
   \multirow{4}{*}{Fake} & 5\% & 1000 & 0 & 53 & 53 & 71.06 ($\pm$ 0.483) &  7.80 ($\pm$ 0.483) \circledE{5}\\ \cline{3-9} 
    &  & 10\% & 1000 & 0 & 112 & 112 & 52.52 ($\pm$ 1.130)& 26.34 ($\pm$ 1.130) \circledE{5} \\ \cline{3-9}
    &  & 20\% & 1000 & 0 & 251 & 251 & 27.55 ($\pm$ 0.636) &  51.31 ($\pm$ 0.636) \circledE{5} \\ \cline{3-9}
    &  & 30\% & 1000 & 0 & 429 & 429 & 25.27 ($\pm$ 0.468) & 53.59 ($\pm$ 0.468)  \circledE{5} \\ \cline{2-9} 

  & \multirow{4}{*}{\vtop{\hbox{\strut Hybrid} \hbox{\strut comp: 0.1\%}}} & 5\% & 999 & 1 & 52 & 100 & 69.73 ($\pm$ 0.773) & 9.13 ($\pm$ 0.773) \circledD{4} \\ \cline{3-9}   
     &  & 10\% & 999 & 1 &  110 & 111 & 49.46 ($\pm$ 0.769) & 29.40 ($\pm$ 0.769) \circledD{4} \\ \cline{3-9}
   &  & 20\% & 999 & 1 &  249 & 250 & 22.05 ($\pm$ 2.498) & 56.81 ($\pm$ 2.498) \circledD{4}  \\ \cline{3-9}
   &  & 30\% & 999 & 1 &  428 & 429 & 10.02 ($\pm$ 3.481) & 68.84 ($\pm$ 3.481) \circledD{4}  \\ \cline{2-9} 

    & \multirow{4}{*}{\vtop{\hbox{\strut Hybrid} \hbox{\strut comp: 0.3\%}}} & 5\% & 997 & 3 & 50 & 300 & 67.70 ($\pm$ 0.699) & 11.16 ($\pm$ 0.699) \circledC{3} \\ \cline{3-9}   
     &  & 10\% & 997 & 3 & 108  & 300 & 46.22 ($\pm$ 1.288) & 32.64 ($\pm$ 1.288) \circledC{3}  \\ \cline{3-9}
     &  & 20\% & 997 & 3 &  247 & 300 & 12.02 ($\pm$ 2.349) & 66.84 ($\pm$ 2.349) \circledC{3}  \\ \cline{3-9}
     &  & 30\% & 997 & 3 &  425 & 428 & 10.00 ($\pm$ 0.000) & 68.86 ($\pm$ 0.000) \circledC{3}  \\ \cline{2-9} 

    & \multirow{4}{*}{\vtop{\hbox{\strut Hybrid} \hbox{\strut comp: 0.5\%}} } & 5\% & 995 & 5 & 48 & 500 & 65.57 ($\pm$ 1.279) & 13.29 ($\pm$ 1.279) \circledB{2} \\ \cline{3-9}    
    &  & 10\% & 995 & 5 &  106 & 500 & 44.79 ($\pm$ 2.152) & 34.07 ($\pm$ 2.152) \circledB{2} \\ \cline{3-9}
    &  & 20\% & 995 & 5 &  244 & 500 & 10.00 ($\pm$ 2.506) &  68.86 ($\pm$ 2.506) \circledB{2} \\ \cline{3-9}
    &  & 30\% & 995 & 5 &  422 & 500 & 10.00 ($\pm$ 0.000) &  68.86 ($\pm$ 0.000) \circledB{2} \\ \cline{2-9}

    & \multirow{4}{*}{Comp} & 5\% & 950 &  50 & 0 & 5,000 & 64.45 ($\pm$ 0.018) & 14.41 ($\pm$ 0.018) \circledA{1} \\ \cline{3-9}    
     &  & 10\% & 900 & 100 & 0 & 10,000 & 41.73 ($\pm$ 2.201) & 37.13 ($\pm$ 2.201) \circledA{1} \\ \cline{3-9}
     &  & 20\% & 800 & 200 & 0 & 20,000 & 10.00 ($\pm$ 0.000) & 68.86 ($\pm$ 0.000) \circledA{1} \\ \cline{3-9}
    &  & 30\% & 700 & 300 & 0 & 30,000 & 10.00 ($\pm$ 0.000) & 68.86 ($\pm$ 0.000) \circledA{1}  \\ \cline{1-9}
   \hline 
\end{tabular}}
\end{table*}

\begin{table*}[hbt!]
\caption{Attack impact ($I_{\theta}$) and maximum test accuracy ($A_{\theta}^M$) of the Norm-Bounding for training on FEMNIST distributed over 3400 initial clients in the presence of different adversaries.} \label{tab:FEMNIST_normb}
\centering
\footnotesize
\resizebox{\textwidth}{!}{
\begin{tabular} {|c|c|c|c|c|c|c||c|c|}
  \hline
   AGR & Attack Type &  \parbox{1.2cm}{Malicious Rate} &  \parbox{1cm}{Number of Benign Clients} & \parbox{1cm}{Number of Compromised Clients}  & \parbox{1cm}{Number of \\Injected Fake Clients \\}  & \parbox{1.1cm}{Attack Cost (\$)} & Accuracy (\%) & Attack Impact (\%) \\ \hline 

     \multirow{20}{*}{ \parbox{1cm}{ Norm Bounding ($\tau=2.0$) (No attack acc = 87.49\%)}}& \multirow{4}{*}{Fake} &  5\% & 3400 & 0 &  179 & 179 & 86.05 ($\pm$ 0.110)  &  1.44 ($\pm$ 0.110) \circledE{5} \\ \cline{3-9}  
    &  & 10\% & 3400 & 0 &  378 & 378 & 83.77 ($\pm$ 0.272) & 3.72 ($\pm$ 0.272) \circledE{5} \\ \cline{3-9}
     &  & 20\% & 3400 & 0 &  850 & 850 & 79.95 ($\pm$ 0.410) & 7.54 ($\pm$ 0.410) \circledE{5} \\ \cline{3-9}
     &  & 30\% & 3400 & 0 &  1458 & 1,458 & 70.45 ($\pm$ 0.470) & 17.04 ($\pm$ 0.470) \circledE{5} \\ \cline{2-9} 

& \multirow{4}{*}{\vtop{\hbox{\strut Hybrid} \hbox{\strut comp: 0.1\%}}} & 5\% & 3396 & 4 & 175 & 400 & 85.65 ($\pm$ 0.188)  &  1.84 ($\pm$ 0.188) \circledC{3} \\ \cline{3-9}   
    &  & 10\% & 3396 & 4 & 374 & 400 & 83.40 ($\pm$ 0.213) & 4.09 ($\pm$ 0.213) \circledD{4} \\ \cline{3-9}
     &  & 20\% & 3396 & 4  & 845 & 849 & 78.99 ($\pm$ 0.487) & 8.50 ($\pm$ 0.487) \circledD{4} \\ \cline{3-9}
    &  & 30\% & 3396 & 4 & 1452 & 1,456 & 70.35 ($\pm$ 0.789) & 17.14 ($\pm$ 0.789) \circledD{4} \\ \cline{2-9} 

    & \multirow{4}{*}{\vtop{\hbox{\strut Hybrid} \hbox{\strut comp: 0.3\%}}} & 5\% & 3389 & 11 &  168 & 1,100  & 85.70 ($\pm$ 0.119)  &  1.79 ($\pm$ 0.119) \circledD{4} \\ \cline{3-9}    
     &  & 10\% & 3389 & 11 & 366 & 1,100 & 83.11 ($\pm$ 0.261) & 4.38 ($\pm$ 0.261) \circledC{3} \\ \cline{3-9}
     &  & 20\% & 3389 & 11 & 837 & 1,100 & 77.74 ($\pm$ 0.614) & 9.75 ($\pm$ 0.614) \circledC{3} \\ \cline{3-9}
     &  & 30\% & 3389 & 11 & 1442 & 1,453 & 70.06 ($\pm$ 2.029) & 17.43 ($\pm$ 2.029) \circledC{3}  \\ \cline{2-9} 

    & \multirow{4}{*}{\vtop{\hbox{\strut Hybrid} \hbox{\strut comp: 0.5\%}}} & 5\% & 3383 & 17 &  162 & 1,700 & 85.56 ($\pm$ 0.143) &  1.93 ($\pm$ 0.143) \circledB{2}\\ \cline{3-9}    
     &  & 10\% & 3383 & 17 & 359 & 1,700 & 82.90 ($\pm$ 0.259) & 4.59 ($\pm$ 0.259) \circledB{2} \\ \cline{3-9}
   &  & 20\% & 3383 & 17 & 829 & 1,700 & 77.72 ($\pm$ 0.395) & 9.77 ($\pm$ 0.395) \circledB{2} \\ \cline{3-9}
     &  & 30\% & 3383 & 17 & 1433 & 1,700 & 69.88 ($\pm$ 0.870) & 17.61 ($\pm$ 0.870) \circledB{2} \\ \cline{2-9}

  & \multirow{4}{*}{Comp} & 5\% & 3230 & 170 & 0 & 17,000 & 84.46 ($\pm$ 0.094)  &  3.03 ($\pm$ 0.094) \circledA{1}\\ \cline{3-9}    
     &  & 10\% & 3060 & 340 & 0 & 34,000 & 81.50 ($\pm$ 0.143) & 5.99 ($\pm$ 0.143) \circledA{1} \\ \cline{3-9}
    &  & 20\% & 2720 & 680 & 0 & 68,000 & 73.28 ($\pm$ 1.332) & 14.21 ($\pm$ 1.332) \circledA{1} \\ \cline{3-9}
     &  & 30\% & 2380 & 1020 & 0 & 102,000 & 52.91 ($\pm$ 1.611) & 34.58 ($\pm$ 1.611) \circledA{1} \\ \cline{1-9} \hline

     \multirow{20}{*}{ \parbox{1cm}{Norm Bounding ($\tau=0.5$) (No attack acc = 86.35\%)}}& \multirow{4}{*}{Fake} & 5\% & 3400 & 0 &  179 & 179 & 85.62 ($\pm$ 0.064)  & 0.73 ($\pm$ 0.064)  \circledE{5} \\ \cline{3-9}    
     &  & 10\% & 3400 & 0 &  378 & 378 & 83.90 ($\pm$ 0.159) & 2.45 ($\pm$ 0.159) \circledE{5} \\ \cline{3-9}
     &  & 20\% & 3400 & 0 &  850 & 850 & 80.33 ($\pm$ 0.315) & 6.02 ($\pm$ 0.315) \circledE{5} \\ \cline{3-9}
     &  & 30\% & 3400 & 0 &  1458 & 1,458 & 71.62 ($\pm$ 1.213) & 14.73 ($\pm$ 1.213) \circledE{5} \\ \cline{2-9} 

& \multirow{4}{*}{\vtop{\hbox{\strut Hybrid} \hbox{\strut comp: 0.1\%}}} & 5\% & 3396 & 4 & 175 & 400 & 85.46 ($\pm$ 0.170)  &  0.89 ($\pm$ 0.170) \circledD{4}\\ \cline{3-9}    
    &  & 10\% & 3396 & 4 & 374 & 400 & 83.58 ($\pm$ 0.201) & 2.77 ($\pm$ 0.201) \circledD{4} \\ \cline{3-9}
     &  & 20\% & 3396 & 4  & 845 & 849 & 79.59 ($\pm$ 0.192) & 6.76 ($\pm$ 0.192) \circledD{4} \\ \cline{3-9}
     &  & 30\% & 3396 & 4 & 1452 & 1,456 & 69.88 ($\pm$ 0.772) & 16.47 ($\pm$ 0.772) \circledD{4} \\ \cline{2-9} 

    & \multirow{4}{*}{\vtop{\hbox{\strut Hybrid} \hbox{\strut comp: 0.3\%}}} & 5\% & 3389 & 11 &  168 & 1,100 & 85.35 ($\pm$ 0.115)  & 1.00 ($\pm$ 0.115)   \circledC{3} \\ \cline{3-9}   
    &  & 10\% & 3389 & 11 & 366 & 1,100 & 83.55 ($\pm$ 0.160) & 2.80 ($\pm$ 0.160) \circledC{3} \\ \cline{3-9}
    &  & 20\% & 3389 & 11 & 837 & 1,100 & 78.88 ($\pm$ 0.177) & 7.47 ($\pm$ 0.177) \circledC{3} \\ \cline{3-9}
    &  & 30\% & 3389 & 11 & 1442 & 1,453 & 69.73 ($\pm$ 1.081) & 16.62 ($\pm$ 1.081) \circledC{3} \\ \cline{2-9} 

    & \multirow{4}{*}{\vtop{\hbox{\strut Hybrid} \hbox{\strut comp: 0.5\%}}} & 5\% & 3383 & 17 &  162 & 1,700 & 85.14 ($\pm$ 0.154)  &  1.21 ($\pm$ 0.154) \circledB{2}\\ \cline{3-9} 
     & & 10\% & 3383 & 17 & 359 & 1,700 & 82.82 ($\pm$ 0.193) & 3.53 ($\pm$ 0.193) \circledB{2} \\ \cline{3-9}
     &  & 20\% & 3383 & 17 & 829 & 1,700 & 78.74 ($\pm$ 0.257) & 7.61 ($\pm$ 0.257) \circledB{2} \\ \cline{3-9}
    &  & 30\% & 3383 & 17 & 1433 & 1,700 & 69.56 ($\pm$ 0.721) & 16.79 ($\pm$ 0.721) \circledB{2} \\ \cline{2-9}  

    & \multirow{4}{*}{Comp} & 5\% & 3230 & 170 & 0 & 17,000 & 84.01 ($\pm$ 0.153) &  2.34 ($\pm$ 0.153)  \circledA{1}\\ \cline{3-9}   
     &  & 10\% & 3060 & 340 & 0 & 34,000 & 81.46 ($\pm$ 0.299) & 4.89 ($\pm$ 0.299) \circledA{1} \\ \cline{3-9}
     &  & 20\% & 2720 & 680 & 0 & 68,000 & 74.47 ($\pm$ 0.745) & 11.88 ($\pm$ 0.745) \circledA{1} \\ \cline{3-9}
     &  & 30\% & 2380 & 1020 & 0 & 102,000 & 56.43 ($\pm$ 2.645) & 29.92 ($\pm$ 2.645) \circledA{1} \\ 
   \hline 
\end{tabular}}
\end{table*}

\end{document}